\documentclass{article}

     \PassOptionsToPackage{numbers, compress}{natbib}
\usepackage[final]{neurips_2021}

\newcommand{\ignore}[1]{}
\usepackage{dsfont}

\usepackage[T1]{fontenc}
\usepackage{amsmath,amssymb,amsfonts,mathrsfs,bm}% Typical maths resource packages
\usepackage{mathtools}
\usepackage{amsthm}
\usepackage{nicefrac}
\usepackage[shortlabels]{enumitem}
\usepackage{graphicx}
\usepackage{epstopdf}
\usepackage{url}
\usepackage{colortbl}
\usepackage{booktabs}
\usepackage{multirow}
\usepackage[table,dvipsnames]{xcolor}
\usepackage[normalem]{ulem}
\usepackage{xparse}

\usepackage{array}
\newcolumntype{L}[1]{>{\raggedright\let\newline\\\arraybackslash\hspace{0pt}}m{#1}}
\newcolumntype{C}[1]{>{\centering\let\newline\\\arraybackslash\hspace{0pt}}m{#1}}
\newcolumntype{R}[1]{>{\raggedleft\let\newline\\\arraybackslash\hspace{0pt}}m{#1}}

\makeatletter
\let\MYcaption\@makecaption
\makeatother
\usepackage[font=footnotesize]{subcaption}
\makeatletter
\let\@makecaption\MYcaption
\makeatother



\usepackage{glossaries}

\makeatletter
\let\oldgls\gls
\let\oldglspl\glspl

\newcommand\fussy@ifnextchar[3]{%
  \let\reserved@d=#1%
  \def\reserved@a{#2}%
  \def\reserved@b{#3}%
  \futurelet\@let@token\fussy@ifnch}
\def\fussy@ifnch{%
  \ifx\@let@token\reserved@d
    \let\reserved@c\reserved@a 
  \else
    \let\reserved@c\reserved@b
  \fi
 \reserved@c}

\renewcommand{\gls}[1]{%
  \oldgls{#1}\fussy@ifnextchar.{\@checkperiod}{\@}}
\renewcommand{\glspl}[1]{%
  \oldglspl{#1}\fussy@ifnextchar.{\@checkperiod}{\@}}

\newcommand{\@checkperiod}[1]{%
  \ifnum\sfcode`\.=\spacefactor\else#1\fi
}
\makeatother

\newacronym{wrt}{w.r.t.}{with respect to}
\newacronym{RHS}{R.H.S.}{right-hand side}
\newacronym{LHS}{L.H.S.}{left-hand side}
\newacronym{iid}{i.i.d.}{independent and identically distributed}
\usepackage{float}

\ifx\notloadhyperref\undefined
	\ifx\loadbibentry\undefined
		\usepackage[hidelinks,hypertexnames=false]{hyperref} 
	\else
		\usepackage{bibentry}
		\makeatletter\let\saved@bibitem\@bibitem\makeatother
		\usepackage[hidelinks,hypertexnames=false]{hyperref}
		\makeatletter\let\@bibitem\saved@bibitem\makeatother
	\fi
\else
	\ifx\loadbibentry\undefined
		\relax
	\else
		\usepackage{bibentry}
	\fi
\fi

\usepackage[capitalize]{cleveref}
\crefname{equation}{}{}
\Crefname{equation}{}{}
\crefname{claim}{claim}{claims}
\crefname{step}{step}{steps}
\crefname{line}{line}{lines}
\crefname{condition}{condition}{conditions}
\crefname{dmath}{}{}
\crefname{dseries}{}{}
\crefname{dgroup}{}{}

\crefname{Problem}{Problem}{Problems}
\crefformat{Problem}{Problem~(#2#1#3)}
\crefrangeformat{Problem}{Problems~(#3#1#4) to~(#5#2#6)}

\crefname{Theorem}{Theorem}{Theorems}
\crefname{Corollary}{Corollary}{Corollaries}
\crefname{Proposition}{Proposition}{Propositions}
\crefname{Lemma}{Lemma}{Lemmas}
\crefname{Definition}{Definition}{Definitions}
\crefname{Example}{Example}{Examples}
\crefname{Assumption}{Assumption}{Assumptions}
\crefname{Remark}{Remark}{Remarks}
\crefname{Rem}{Remark}{Remarks}
\crefname{remarks}{Remarks}{Remarks}
\crefname{Appendix}{Appendix}{Appendices}
\crefname{Supplement}{Supplement}{Supplements}
\crefname{Exercise}{Exercise}{Exercises}
\crefname{Theorem_A}{Theorem}{Theorems}
\crefname{Corollary_A}{Corollary}{Corollaries}
\crefname{Proposition_A}{Proposition}{Propositions}
\crefname{Lemma_A}{Lemma}{Lemmas}
\crefname{Definition_A}{Definition}{Definitions}

\usepackage{crossreftools}
\ifx\notloadhyperref\undefined
\else
	\relax
\fi

\usepackage{algorithm,algorithmic}

\ifx\loadbreqn\undefined
	\relax
\else
	\usepackage{breqn} 
\fi

\ifx\renewtheorem\undefined
\ifx\useTheoremCounter\undefined
\newtheorem{Theorem}{Theorem}
\newtheorem{Corollary}{Corollary}
\newtheorem{Proposition}{Proposition}
\newtheorem{Lemma}{Lemma}
\else
\newtheorem{Theorem}{Theorem}
\newtheorem{Corollary}[theorem]{Corollary}

\fi

\newtheorem{Definition}{Definition}

\newtheorem{Assumption}{Assumption}

\fi

\newcommand{\nn}{\nonumber\\ }

\theoremstyle{remark}

\theoremstyle{plain}

\newcommand{\qednew}{\nobreak \ifvmode \relax \else
      \ifdim\lastskip<1.5em \hskip-\lastskip
      \hskip1.5em plus0em minus0.5em \fi \nobreak
      \vrule height0.75em width0.5em depth0.25em\fi}

\NewDocumentCommand{\movedownsub}{e{^_}}{%
  \IfNoValueTF{#1}{%
    \IfNoValueF{#2}{^{}}% neither ^ nor _, do nothing; if no ^ but _, add ^{}
  }{%
    ^{#1}% add superscript if present
  }%
  \IfNoValueF{#2}{_{#2}}% add subscript if present
}

\newcommand{\Real}{\mathbb{R}}

\newcommand{\Complex}{\mathbb{C}}

\newcommand{\calL}{\mathcal{L}}

\newcommand{\bA}{\mathbf{A}}

\newcommand{\bB}{\mathbf{B}}

\newcommand{\bC}{\mathbf{C}}

\newcommand{\bI}{\mathbf{I}}

\newcommand{\bQ}{\mathbf{Q}}

\newcommand{\bU}{\mathbf{U}}
\newcommand{\bv}{\mathbf{v}}
\newcommand{\bV}{\mathbf{V}}

\newcommand{\bz}{\mathbf{z}}

\DeclareSymbolFont{bsfletters}{OT1}{cmss}{bx}{n}
\DeclareSymbolFont{ssfletters}{OT1}{cmss}{m}{n}
\DeclareMathSymbol{\bsfGamma}{0}{bsfletters}{'000}
\DeclareMathSymbol{\ssfGamma}{0}{ssfletters}{'000}
\DeclareMathSymbol{\bsfDelta}{0}{bsfletters}{'001}
\DeclareMathSymbol{\ssfDelta}{0}{ssfletters}{'001}
\DeclareMathSymbol{\bsfTheta}{0}{bsfletters}{'002}
\DeclareMathSymbol{\ssfTheta}{0}{ssfletters}{'002}
\DeclareMathSymbol{\bsfLambda}{0}{bsfletters}{'003}
\DeclareMathSymbol{\ssfLambda}{0}{ssfletters}{'003}
\DeclareMathSymbol{\bsfXi}{0}{bsfletters}{'004}
\DeclareMathSymbol{\ssfXi}{0}{ssfletters}{'004}
\DeclareMathSymbol{\bsfPi}{0}{bsfletters}{'005}
\DeclareMathSymbol{\ssfPi}{0}{ssfletters}{'005}
\DeclareMathSymbol{\bsfSigma}{0}{bsfletters}{'006}
\DeclareMathSymbol{\ssfSigma}{0}{ssfletters}{'006}
\DeclareMathSymbol{\bsfUpsilon}{0}{bsfletters}{'007}
\DeclareMathSymbol{\ssfUpsilon}{0}{ssfletters}{'007}
\DeclareMathSymbol{\bsfPhi}{0}{bsfletters}{'010}
\DeclareMathSymbol{\ssfPhi}{0}{ssfletters}{'010}
\DeclareMathSymbol{\bsfPsi}{0}{bsfletters}{'011}
\DeclareMathSymbol{\ssfPsi}{0}{ssfletters}{'011}
\DeclareMathSymbol{\bsfOmega}{0}{bsfletters}{'012}
\DeclareMathSymbol{\ssfOmega}{0}{ssfletters}{'012}

\newcommand{\btheta}{\bm{\theta}}

\newcommand{\bphi}{\bm{\phi}}

\newcommand{\bSigma	}{\bm{\Sigma}}

\DeclareMathOperator{\st}{s.t.\ }
\DeclarePairedDelimiter\abs{\lvert}{\rvert}

\DeclarePairedDelimiterX\ip[2]{\langle}{\rangle}{#1,#2}
\DeclarePairedDelimiterX\norm[1]{\lVert}{\rVert}{#1}
\DeclarePairedDelimiterXPP\col[1]{\operatorname{col}}{\{}{\}}{}{#1} % column vector
\DeclarePairedDelimiterXPP\row[1]{\operatorname{row}}{\{}{\}}{}{#1} % row vector
\DeclarePairedDelimiterXPP\erf[1]{\operatorname{erf}}{(}{)}{}{#1}
\DeclarePairedDelimiterXPP\erfc[1]{\operatorname{erfc}}{(}{)}{}{#1}
\DeclarePairedDelimiterXPP\op[2]{\operatorname{#1}}{(}{)}{}{#2} % general operator

\newcommand{\T}{^{\intercal}}% transpose notation
\newcommand{\ud}{\,\mathrm{d}} % for integrals like \int f(x) \ud x
\newcommand{\ddfrac}[2]{{\dfrac{\mathrm{d} {#1}}{\mathrm{d} {#2}}}}

\newcommand{\tc}[1]{^{(#1)}}

\DeclarePairedDelimiterX\Set[2]\{\}{%

#2
}
\DeclarePairedDelimiterX\Setc[1]\{\}{%

#1
}

\NewDocumentCommand\set{s o m}{%
	\IfBooleanTF#1%
	{\IfValueTF{#2}{\Set*{#2}{#3}}{\Setc*{#3}}}%
	{\IfValueTF{#2}{\Set{#2}{#3}}{\Setc{#3}}}%
}

\NewDocumentCommand{\evalat}{s O{\big} m m}{%
  \IfBooleanTF{#1}
   {{\left. #3 \right|_{#4}}}
   {{#3#2|_{#4}}}%
}

\NewDocumentCommand \ifcond {m m} {%
	{#1} %
	\IfValueT{#2}{\, \middle|\, {#2}}%
}

\DeclareDocumentCommand \P {e{_} g >{\SplitArgument{ 1 }{ @| }}d() g } {%
	\mathbb{P}%
	\IfValueTF{#1}{_{#1}}
		{\IfValueT{#2}{_{#2}}}%
	\IfValueT{#3}{\left(\ifcond#3}%
	\IfValueT{#4}{\, \middle|\, {#4}}%
	\IfValueT{#3}{\right)}%
}

\DeclareDocumentCommand \E {e{_} g >{\SplitArgument{ 1 }{ @| }}o g } {%
	\mathbb{E}%
	\IfValueTF{#1}{_{#1}}
		{\IfValueT{#2}{_{#2}}}%
	\IfValueT{#3}{\left[\ifcond#3}%
	\IfValueT{#4}{\, \middle|\, {#4}}%
	\IfValueT{#3}{\right]}%
}

\let\oldforall\forall
\renewcommand{\forall}{\oldforall \, }

\let\oldexist\exists
\renewcommand{\exists}{\oldexist \: }

\graphicspath{{./Figures/}} 
\newcommand{\includeCroppedPdf}[2][]{%
    \IfFileExists{./Figures/#2-crop.pdf}{}{%
        \immediate\write18{pdfcrop ./Figures/#2 ./Figures/#2-crop.pdf}}%
    \includegraphics[#1]{./Figures/#2-crop.pdf}}

\definecolor{gray90}{gray}{0.9}

\ifx\nohighlights\undefined

	\newcommand{\msout}[1]{\text{\color{green} \sout{\ensuremath{#1}}}}
	\newcommand{\del}[1]{{\color{green}\ifmmode \msout{#1}\else\sout{#1}\fi}}
\else

	\newcommand{\msout}[1]{#1}
	\newcommand{\del}[1]{#1}
\fi

\newcommand{\hhide}[1]{}
\ifx\diagnoselabel\undefined
	\relax
\else
	\makeatletter
	 \def\@testdef #1#2#3{%
		 \def\reserved@a{#3}\expandafter \ifx \csname #1@#2\endcsname
		\reserved@a  \else
	 \typeout{^^Jlabel #2 changed:^^J%
	 \meaning\reserved@a^^J%
	 \expandafter\meaning\csname #1@#2\endcsname^^J}%
	 \@tempswatrue \fi}
	\makeatother
\fi

\makeatletter
\newcommand*{\addFileDependency}[1]{% argument=file name and extension
  \typeout{(#1)}
  \@addtofilelist{#1}
  \IfFileExists{#1}{}{\typeout{No file #1.}}
}
\makeatother

\usepackage{tikz}
\usetikzlibrary{shapes.geometric, arrows}
\newcommand{\tb}[1]{\textbf{#1}}
\usepackage[utf8]{inputenc} % allow utf-8 input
\usepackage{microtype}      % microtypography

\title{Stable Neural ODE with Lyapunov-Stable Equilibrium Points for Defending Against Adversarial Attacks}
\author{%
  Qiyu~Kang\thanks{First two authors contributed equally to this work.}\\
  Continental-NTU Corporate Lab\\
  Nanyang Technological University\\
  50 Nanyang Avenue, 639798, Singapore \\
   \texttt{kang0080@e.ntu.edu.sg} \\
   \And
   Yang~Song\footnotemark[1]\\
  School of Electrical and Electronic Engineering\\
  Nanyang Technological University\\
   50 Nanyang Avenue, 639798, Singapore \\
   \texttt{songy@ntu.edu.sg} \\
   \And 
   Qinxu~Ding\\
   School of Business\\
 Singapore University of Social Sciences\\
  463 Clementi Road, 599494, Singapore\\
   \texttt{qinxuding@suss.edu.sg} \\
   \And
   Wee~Peng~Tay\\
   School of Electrical and Electronic Engineering\\
  Nanyang Technological University\\
   50 Nanyang Avenue, 639798, Singapore \\
   \texttt{wptay@ntu.edu.sg} \\
}

\begin{document}

\maketitle

\begin{abstract}

Deep neural networks (DNNs) are well-known to be vulnerable to adversarial attacks, where malicious human-imperceptible perturbations are included in the input to the deep network to fool it into making a wrong classification. Recent studies have demonstrated that neural Ordinary Differential Equations (ODEs) are intrinsically more robust against adversarial attacks compared to vanilla DNNs. In this work, we propose a stable neural ODE with Lyapunov-stable equilibrium points for defending against adversarial attacks (SODEF). By ensuring that the equilibrium points of the ODE solution used as part of SODEF is Lyapunov-stable, the ODE solution for an input with a small perturbation converges to the same solution as the unperturbed input. We provide theoretical results that give insights into the stability of SODEF as well as the choice of regularizers to ensure its stability. Our analysis suggests that our proposed regularizers force the extracted feature points to be within a neighborhood of the Lyapunov-stable equilibrium points of the ODE. SODEF is compatible with many defense methods and can be applied to any neural network's final regressor layer to enhance its stability against adversarial attacks. 
\end{abstract}

\section{Introduction}
\label{sect:intro}
Although deep learning has found successful applications in many tasks such as image classification \cite{KrizhevskyNIPS2012, LecunPIEEE1998}, speech recognition \cite{HintonSPM2012}, and natural language processing \cite{AndorACL2016}, the vulnerability of deep learning to adversarial attacks (e.g., see \cite{SzegedyICLR2013}) has limited its real-world applications due to performance and safety concerns in critical applications. Inputs corrupted with human-imperceptible perturbations can easily fool many vanilla deep neural networks (DNNs) into mis-classifying them and thus significantly impact their performance. 

Recent studies \cite{yan2019robustness,haber2017stable,liu2020does} have applied neural Ordinary Differential Equations (ODEs) \cite{chen2018neural} to defend against adversarial attacks. Some works like \cite{yan2019robustness} have revealed interesting intrinsic properties of ODEs that make them more stable than conventional convolutional neural networks (CNNs). The paper \cite{yan2019robustness} proposes a time-invariant steady neural ODE (TisODE) using the property that the integral curves from a ODE solution starting from different initial points (inputs) do not intersect and always preserve uniqueness in the solution function space. However, this does not guarantee that small perturbations of the initial point lead to small perturbations of the integral curve output at a later time $T$. The authors thus proposed a regularizer to limit the evolution of the curves by forcing the integrand to be close to zero. However, neither the non-intersecting property nor the steady-state constraint used in TisODE can guarantee robustness against input perturbations since these constraints do not ensure that the inputs are within a neighborhood of Lyapunov-stable equilibrium points. An example is an ODE that serves as an identity mapping is not robust to input perturbations but satisfies all the constraints proposed in \cite{yan2019robustness}. 

In this paper, our objective is to design a neural ODE such that the features extracted are within a neighborhood of the Lyapunov-stable equilibrium points of the ODE. We first develop a diversity promoting technique applied in the final fully connected (FC) layer to improve the ODE's stability and analyze the reasons why. We then propose a stable neural ODE with Lyapunov-stable equilibrium points to eliminate the effects of perturbations in the input. From linear control theory \cite{chen1999linear}, a linear time-invariant system $\ud\bz(t)/\ud t= \bA \bz(t)$, where $\bA$ is a constant matrix, is exponentially stable if all eigenvalues of $\bA$ have negative real parts. Specifically, we propose to force the Jacobian matrix of the ODE used in the neural ODE to have eigenvalues with negative real parts. Instead of directly imposing constraints on the eigenvalues of the matrix, which lead to high computational complexity when the Jacobian matrix is large, we instead add constraints to the matrix elements to implicitly force the real parts of its eigenvalues to be negative.

Our main contributions are summarized as follows:
\begin{enumerate}
\item Based on the concept of Lyapunov-stable equilibrium points, we propose a simple yet effective technique to improve the robustness of neural ODE networks by fixing the final FC layer to be a matrix whose rows have unit norm and such that the maximum cosine similarity between any two rows is minimized. Such a FC layer can be constructed off-line.
\item We propose a stable neural ODE for deFending against adversarial attacks (SODEF) to suppress the input perturbations. We derive an optimization formulation for SODEF to force the extracted feature points to be within a neighborhood of the Lyapunov-stable equilibrium points of the SODEF ODE. We provide sufficient conditions for learning a robust feature representation under SODEF.
\item We test SODEF on several widely used datasets MNIST \cite{LeCunMNIST}, CIFAR-10 and CIFAR-100 \cite{KriTR2009} under well-known adversarial attacks. We demonstrate that SODEF is robust against adversarial white-box attacks with improvement in classification accuracy of adversarial examples under PGD attack \cite{MadryICLR2018} of up to $44.02\%$,  $52.54\%$ and $18.91\%$ percentage points compared to another current state-of-the-art neural ODE network TisODE \cite{yan2019robustness} on MNIST, CIFAR-10 and CIFAR-100, respectively. Similar improvements in classification accuracy of adversarial examples of up to $43.69\%$, $52.38\%$ and $18.99\%$ percentage points compared to ODE net \cite{chen2018neural} are also obtained.
\end{enumerate} 
The rest of this paper is organized as follows. We provide essential preliminaries on neural ODE and its stability analysis in \cref{sect:preliminaries}. In \cref{sect:SODEF}, we present SODEF model architecture and its training method. We show how to maximize the distance between stable equilibrium points of neural ODEs. We propose an optimization and present theoretical results on its stability properties. We summarize experimental results in \cref{sect:exper} and conclude the paper in \cref{sect:conc}. The proofs for all lemmas and theorems proposed in this paper are given in the supplementary material. We also refer interested readers to the supplementary material for a more detailed account of related works \cite{Hab2018stable,huang2020adversarial,Li2020IEskip,yan2019robustness} and some popular adversarial attacks \cite{GoodICLR2015,MadryICLR2018} that are used to verify the robustness of our proposed SODEF. In the paper, we use lowercase boldface characters like $\bz$ to denote vectors in $\Real^n$, capital boldface characters like $\bA$ to denote matrices in $\Real^{n \times n}$, and normal characters like $z$ to denote scalars except that the notation $(x,y)$ are normal characters reserved to denote the input and label pairs. A vector $\bz\in\Real^n$ is represented as $(\bz^{(1)},\bz^{(2)},\ldots,\bz^{(n)})$. The $(i,j)$-th element of a matrix $\bA$ is $\bA_{ij}$ or $[\bA]_{ij}$. The Jacobian matrix of a function $f: \Real^n\mapsto\Real^n$ evaluated at $\bz$ is denoted as $\nabla f(\bz)$. The set of functions $\Real^n\mapsto\Real^n$ with continuous first derivatives is denoted as $C^1(\Real^n,\Real^n)$.

\section{Preliminaries: Neural ODE and Stability}
\label{sect:preliminaries}
%\begin{tikzpicture}[node distance=2cm]
%\node (input_image) [input] {$x$};
%\node (feature_extractor) [process, right of=input_image, xshift=2cm] {Feature Extractor $h$};
%\node (fc) [process, right of=feature_extractor, xshift=2cm] {FC};
%\node (output_label) [output, right of=fc, xshift=2cm] {$y$};
%\draw [->] (input_image) -- (feature_extractor);
%\draw [->] (feature_extractor) -- (fc);
%\draw [->] (fc) -- (output_label);
%\end{tikzpicture}

%\begin{tikzpicture}[node distance=2cm]
%\node (input_image) [input] {$x$};
%\node (feature_extractor) [process, right of=input_image, xshift=2cm] {Feature Extractor $h$};
%\node (ODE) [process, right of=feature_extractor, xshift=2cm] {ODE Net $f_{\btheta}$};
%\node (output_label) [output, right of=ODE, xshift=2cm] {$y$};
%\draw [->] (input_image) -- (feature_extractor);
%\draw [->] (feature_extractor) -- (ODE);
%\draw [->] (ODE) -- (output_label);
%\end{tikzpicture}

In a neural ODE layer, the relation between the layer input $\bz(0)$ and output $\bz(T)$ is described as the following differential equation: 
\begin{align}
\ddfrac{\bz(t)}{t}=f_{\btheta}(\bz(t), t) \label{eq:NODE}
\end{align}
where $f_{\btheta}:\Real^n \times [0,\infty)\mapsto \Real^n$ denotes the non-linear trainable layers that are parameterized by weights $\btheta$ and $\bz: [0,\infty) \mapsto \Real^n$ represents the $n$-dimensional state of the neural ODE. Neural ODEs are the continuous analog of residual networks where the hidden layers of residual networks can be regarded as discrete-time difference equations $\bz(t+1)=\bz(t) + f_{\btheta}(\bz(t),t)$. In this work, for simplicity, we only consider the time-invariant (autonomous) case $f_{\btheta}(\bz(t), t) = f_{\btheta}(\bz(t))$, where the dynamical system does not explicitly depend on $t$. For such non-linear dynamical systems, the following theorem shows that under mild conditions, its behaviour can be studied via linearization near special points called hyperbolic equilibrium points.

\begin{Theorem}[Hartman–Grobman Theorem \cite{arrowsmith1992dynamical}]\label{thm:linearization}
Consider a system evolving in time with state $\bz(t)\in \Real^{n}$ that satisfies the differential equation  $\ddfrac{\bz(t)}{t}= f(\bz(t))$ for some \ignore{ smooth map} $f\in C^1(\Real^n,\Real^n)$, $f(\bz)=(f^{(1)}(\bz),\ldots,f^{(n)}(\bz))$. Suppose the map has a hyperbolic equilibrium state $\bz^{*}\in\Real ^{n}$, i.e., $f(\bz^{*})=0$ and the Jacobian matrix $\nabla f =[\partial f^{(i)}/\partial \bz^{(j)}]_{i,j=1}^n$ of $f$ evaluated at $\bz=\bz^{*}$ has no eigenvalue with real part equal to zero. Then there exists a neighbourhood $N_{\bz^{*}}$ of the equilibrium point $\bz^{*}$ and a homeomorphism $g:N_{\bz^{*}}\mapsto\Real ^{n}$, such that $g(\bz^{*})=0$ and in the neighbourhood $N_{\bz^{*}}$, the flow of $\ddfrac{\bz(t)}{t}= f(\bz(t))$ is topologically conjugate by the continuous map $\bar{\bz}(t)=g(\bz(t))$ to the flow of its linearization $\ddfrac{\bar{\bz}(t)}{t}= \nabla f(\bz^*) \cdot \bar{\bz}(t)$.
\end{Theorem}

The theorem states that when the Jacobian matrix at the zeros of $f$ has no eigenvalue with zero real part, the behaviour of the original dynamical system can be studied using the simpler linearization of the system around those zeros. We next review some definitions and theorems from linear control theory \cite{chen1999linear}.

\begin{Definition}[Lyapunov Stability \cite{chen1999linear}] %[Definition 5.1] in chen1999linear
The  linear time-invariant system $\ddfrac{\bar{\bz}(t)}{t}= \bA\bar{\bz}(t)$ with constant matrix $\bA$ is marginally stable or stable in the sense of Lyapunov if every finite initial state $\bar{\bz}(0)$ excites a bounded response. It is asymptotically stable if every finite initial state excites a bounded response, which, in addition, approaches $0$ as $t\to\infty$.
\end{Definition}

\begin{Theorem}[Lyapunov Stability Theorem \cite{chen1999linear}]\label{thm:stab_lin} 
a) The equation $\ddfrac{\bar{\bz}(t)}{t}= \bA\bar{\bz}(t))$   is marginally stable if and only if all eigenvalues of A have zero or negative real parts and those with zero real parts are simple roots of the minimal polynomial of $\bA$. b) The equation $\ddfrac{\bar{\bz}(t)}{t}= \bA\bar{\bz}(t)$  is asymptotically stable if and only if all eigenvalues of A have negative real parts.
\end{Theorem}

In \cref{thm:linearization}, we say that a hyperbolic equilibrium point is \emph{Lyapunov-stable} if all eigenvalues of the Jacobian matrix evaluated at it have negative real parts. From \cref{thm:linearization,thm:stab_lin}, we see that a small perturbation around the Lyapunov-stable equilibrium point $\bz(0)$ leads to $\tilde{\bz}(t) \to \bz(0)$ as $t\to \infty$, i.e., $\exists \delta>0$ such that for all $\tilde{\bz}(0)$ with $\norm{\bz(0)-\tilde{\bz}(0)}_2<\delta$, we have $\norm{\tilde{\bz}(t)-\bz(0)}_2\to 0$ as $t\to \infty$, where $\tilde{\bz}(t)$ is the ODE solution for the perturbed input $\tilde{\bz}(0)$. In the context of neural network adversarial attacks, if the malicious perturbations around the ODE input $\bz(0)$ is small, then the output  $\bz(T)$ for large enough $T$ will not be affected significantly by the perturbation. Consequently, the succeeding network layers after the neural ODE layer can still perform well without being affected by the input perturbation. The perturbation weakening phenomenon around Lyapunov-stable equilibrium points works like a noise filter and acts as a defense against adversarial attacks.

We require the following definition and result in our stability analysis.

\begin{Definition}[Strictly diagonally dominant \cite{horn2012matrix}] 
Let $\bA\in \Complex^{n \times n}$ . We say that $\bA$ is strictly diagonally dominant if  $|\bA_{ii}| > \sum_{j\ne i}|\bA_{ij}|$ for all $i = 1,...,n$.
\end{Definition}

\begin{Theorem}[Levy–Desplanques theorem \cite{horn2012matrix}]\label{thm:Taussky}
If $\bA\in \Complex^{n \times n}$ is strictly diagonally dominant and if every main diagonal entry of A is real and negative, then $A$ is non-singular and every eigenvalue of A has negative real part.
\end{Theorem}

\begin{Lemma}\label{lem:exist_f}
Given  $k$ distinct points $\bz_i\in \Real^n$ and matrices $\bA_i\in \Real^{n\times n}$, $i= 1,...,k$, there exists a function $f\in C^1(\Real^n,\Real^n)$ such that $f(\bz_i)=0$ and $\nabla f_{\btheta}(\bz_i) = \bA_i$.
\end{Lemma}

\section{SODEF Architecture}
\label{sect:SODEF}

\begin{figure}
\centering

\tikzset{every picture/.style={line width=0.75pt}} %set default line width to 0.75pt        

\begin{tikzpicture}[x=0.75pt,y=0.75pt,yscale=-1,xscale=1]
%uncomment if require: \path (0,545); %set diagram left start at 0, and has height of 545

%Straight Lines [id:da879193758003253] 
\draw    (40.33,140.17) -- (63,140.47) ;
\draw [shift={(65,140.5)}, rotate = 180.77] [color={rgb, 255:red, 0; green, 0; blue, 0 }  ][line width=0.75]    (10.93,-3.29) .. controls (6.95,-1.4) and (3.31,-0.3) .. (0,0) .. controls (3.31,0.3) and (6.95,1.4) .. (10.93,3.29)   ;
%Rounded Rect [id:dp7572288881061273] 
\draw   (6,133.2) .. controls (6,130.7) and (8.03,128.67) .. (10.53,128.67) -- (35.47,128.67) .. controls (37.97,128.67) and (40,130.7) .. (40,133.2) -- (40,146.8) .. controls (40,149.3) and (37.97,151.33) .. (35.47,151.33) -- (10.53,151.33) .. controls (8.03,151.33) and (6,149.3) .. (6,146.8) -- cycle ;
%Shape: Rectangle [id:dp13421291071056785] 
\draw   (65,119.33) -- (196.5,119.33) -- (196.5,160.67) -- (65,160.67) -- cycle ;
%Shape: Rectangle [id:dp246732414193521] 
\draw  [dash pattern={on 4.5pt off 4.5pt}] (228,120.67) -- (371,120.67) -- (371,162.5) -- (228,162.5) -- cycle ;
%Shape: Rectangle [id:dp29809940727973316] 
\draw   (402.17,121.33) -- (451.17,121.33) -- (451.17,161.33) -- (402.17,161.33) -- cycle ;
%Straight Lines [id:da572889539783221] 
\draw    (197,140.75) -- (227.67,140.67) ;
\draw [shift={(229.67,140.67)}, rotate = 539.85] [color={rgb, 255:red, 0; green, 0; blue, 0 }  ][line width=0.75]    (10.93,-3.29) .. controls (6.95,-1.4) and (3.31,-0.3) .. (0,0) .. controls (3.31,0.3) and (6.95,1.4) .. (10.93,3.29)   ;
%Rounded Rect [id:dp6294962587179136] 
\draw   (483.17,132.03) .. controls (483.17,129.53) and (485.2,127.5) .. (487.7,127.5) -- (512.63,127.5) .. controls (515.14,127.5) and (517.17,129.53) .. (517.17,132.03) -- (517.17,145.63) .. controls (517.17,148.14) and (515.14,150.17) .. (512.63,150.17) -- (487.7,150.17) .. controls (485.2,150.17) and (483.17,148.14) .. (483.17,145.63) -- cycle ;
%Straight Lines [id:da44537392246306595] 
\draw    (370,140.75) -- (400.67,140.67) ;
\draw [shift={(402.67,140.67)}, rotate = 539.85] [color={rgb, 255:red, 0; green, 0; blue, 0 }  ][line width=0.75]    (10.93,-3.29) .. controls (6.95,-1.4) and (3.31,-0.3) .. (0,0) .. controls (3.31,0.3) and (6.95,1.4) .. (10.93,3.29)   ;
%Straight Lines [id:da5420590881542178] 
\draw    (450.5,140.75) -- (481.17,140.67) ;
\draw [shift={(483.17,140.67)}, rotate = 539.85] [color={rgb, 255:red, 0; green, 0; blue, 0 }  ][line width=0.75]    (10.93,-3.29) .. controls (6.95,-1.4) and (3.31,-0.3) .. (0,0) .. controls (3.31,0.3) and (6.95,1.4) .. (10.93,3.29)   ;

% Text Node
\draw (17.53,136.67) node [anchor=north west][inner sep=0.75pt]  [font=\normalsize] [align=left] {$\displaystyle x$};
% Text Node
\draw (494.5,134.17) node [anchor=north west][inner sep=0.75pt]   [align=left] {$\displaystyle y$};
% Text Node
\draw (78,128.33) node [anchor=north west][inner sep=0.75pt]   [align=left] { \\ feature extractor $\displaystyle h_{\bphi}$};
% Text Node
\draw (234.67,123.67) node [anchor=north west][inner sep=0.75pt]   [align=left] {ODE $\ddfrac{\bz(t)}{t} = f_{\btheta}(\bz(t))$};
% Text Node
\draw (410.5,134) node [anchor=north west][inner sep=0.75pt]   [align=left] {FC $\bV$};
% Text Node
\draw (198.5,119.33) node [anchor=north west][inner sep=0.75pt]   [align=left] {$\bz(0)$};
% Text Node
\draw (371.83,119.67) node [anchor=north west][inner sep=0.75pt]   [align=left] {$\bz(T)$};

\end{tikzpicture}

\caption{SODEF model architecture.}\label{fig:structure}
\end{figure}
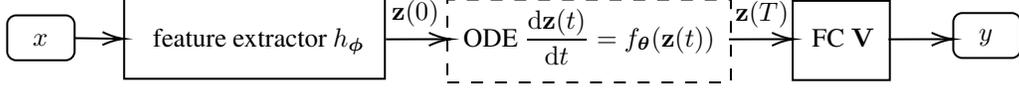

We consider a classification problem with $L$ classes. The proposed SODEF model architecture is shown in \cref{fig:structure}. The input $x\in X$ (e.g., an image) is first passed through a feature extractor $h_{\bphi} : X \mapsto\Real^n$ to obtain an embedding feature representation $\bz(0)$. A neural ODE layer $f_{\btheta }$ follows as a nonlinear feature mapping to stabilize the feature representation output $\bz(0)$ from $h_{\bphi}$. The final FC layer $\bV$ serves as a linear mapping to generate a prediction vector based on the output $\bz(T)$ of the neural ODE layer. The parameters $\bphi,\btheta$ and $\bV$ are parameterized weights for the feature extractor, neural ODE layer and FC layer, respectively. 

We provide motivation and design guidance for the FC layer $\bV$ in \cref{subsect:max_dis}, which attempts to separate Lyapunov-stable equilibrium points implicitly by maximizing the similarity distance between feature representations corresponding to the $L$ different classes. Experimental results demonstrate the advantages of our diversity promoting FC layer in \cref{subsect:max_dis} with comparisons to traditional neural ODEs without diversity promoting.

However, the embedded features after using diversity promoting are not guaranteed to locate near the Lyapunov-stable equilibrium points. In \cref{subsect:SODEF_Formulation}, we formulate an optimization problem to force embedding features to locate near the Lyapunov-stable equilibrium points. We introduce optimization constraints to force the Jacobian matrix of the ODE in our neural ODE layer to have eigenvalues with negative real parts at the Lyapunov-stable equilibrium points. Instead of directly imposing constraints on the eigenvalue of the matrix, which may be computationally complex especially when the matrix is large, we add constraints to the matrix elements instead.

\subsection{Maximizing the Distance between Lyapunov-Stable Equilibrium Points}
\label{subsect:max_dis}

From \cref{sect:preliminaries}, we observe that points in a small neighbourhood of a Lyapunov-stable equilibrium point is robust against adversarial perturbations. We call this neighborhood a \emph{stable neighborhood}. However Lyapunov-stable equilibrium points for different classes may very well locate near each other and therefore each stable neighborhood may be very small, leading to poor adversarial defense. In this section, we propose to add a FC layer after the neural ODE layer given by \cref{eq:NODE} to avoid this scenario. The purpose of the FC layer is to map the output of the neural ODE layer to a feature vector $\bv_l$ if the input $x$ belongs to the class $l=1,\ldots,L$. We design the FC layer so that the cosine similarities between different $\bv_l$'s are minimized.

%{[do we need to say minmax>maxmin, so we actually also get an upper bound if we try to max the minimum of the pairwise distance]}

%{[can we get some thing about $\E d(\bz_i,\bz_j)$, if $\bz_i\in \text{neighbour}(\bv_l)$ and $\bz_i\in \text{neighbour}(\bv_l')$]}

\begin{Lemma}\label{thm:reg_simplex}
Given a set of $k$ unit vectors $\bv_1,\ldots,\bv_k$ in $\Real^{n}$, where $n\ge k$, let $a(\bv_1,\ldots,\bv_k)=\max_{i\neq j} \ \bv_i\T\bv_j$. Then $\min a(\bv_1,\ldots,\bv_k) = 1/(1-k)$, where the minimum is taken over all possible sets of $k$ unit vectors $\bv_1,\ldots,\bv_k$.
\end{Lemma}

\begin{Corollary}\label{cor:reg_simplex}
Consider a $k\times k$ matrix $\bB=[b_{ij}]_{i,j=1}^{k}$ with $b_{ii}=1$ and $b_{ij}=1/(1-k),\ \forall i\neq j$.  Let the eigen decomposition of $\bB$ be $\bB=\bU\bSigma\bU\T$. For any $n\geq k$ and $i=1,\ldots,k$, let $\bv_i$ be the $i$-th column of $\bQ\bSigma^{1/2}\bU\T$, where $\bQ$ is any $n\times k$ matrix such that $\bQ\T\bQ=\bI_k$. Then, $a(\bv_1,\ldots,\bv_k)=\max_{i\neq j} \bv_i\T\bv_j= 1/(1-k)$.
\end{Corollary}

\cref{cor:reg_simplex} suggests a diversity promoting scheme to maximally separate the equilibrium points of the neural ODE layer. The FC layer is represented by an $n\times L$ matrix $\bV=[\bv_1,\ldots,\bv_L]$, where $n$ is the dimension of $\bz(T)$, the output from the neural ODE layer. If $\bz(T)$ is generated from an input from class $l$, it is mapped to $\bv_l$. By minimizing the maximum cosine similarity $a(\bv_1,\ldots,\bv_k)=\max_{i\neq j} \bv_i\T\bv_j$ between the representations from two different classes, we ensure that the output of SODEF is robust to perturbations in the input. \cref{cor:reg_simplex} provides a way to choose the FC layer weights $\bV$. 

%\subsection{Training with fixed last FC}\label{subsec:fixfcexp}

\begin{figure}[!htb]
\centering
    \includegraphics[width=0.32\textwidth]{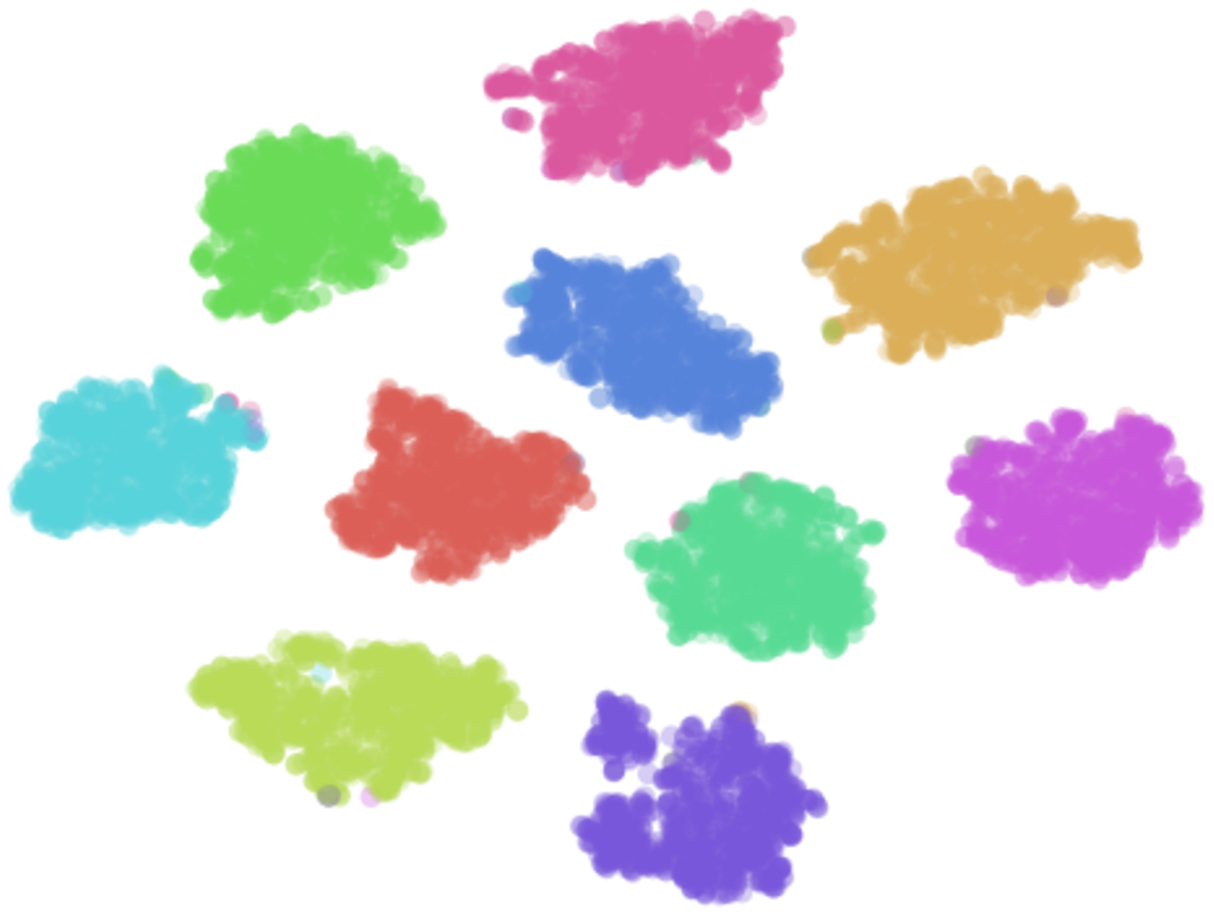}
    \includegraphics[width=0.32\textwidth]{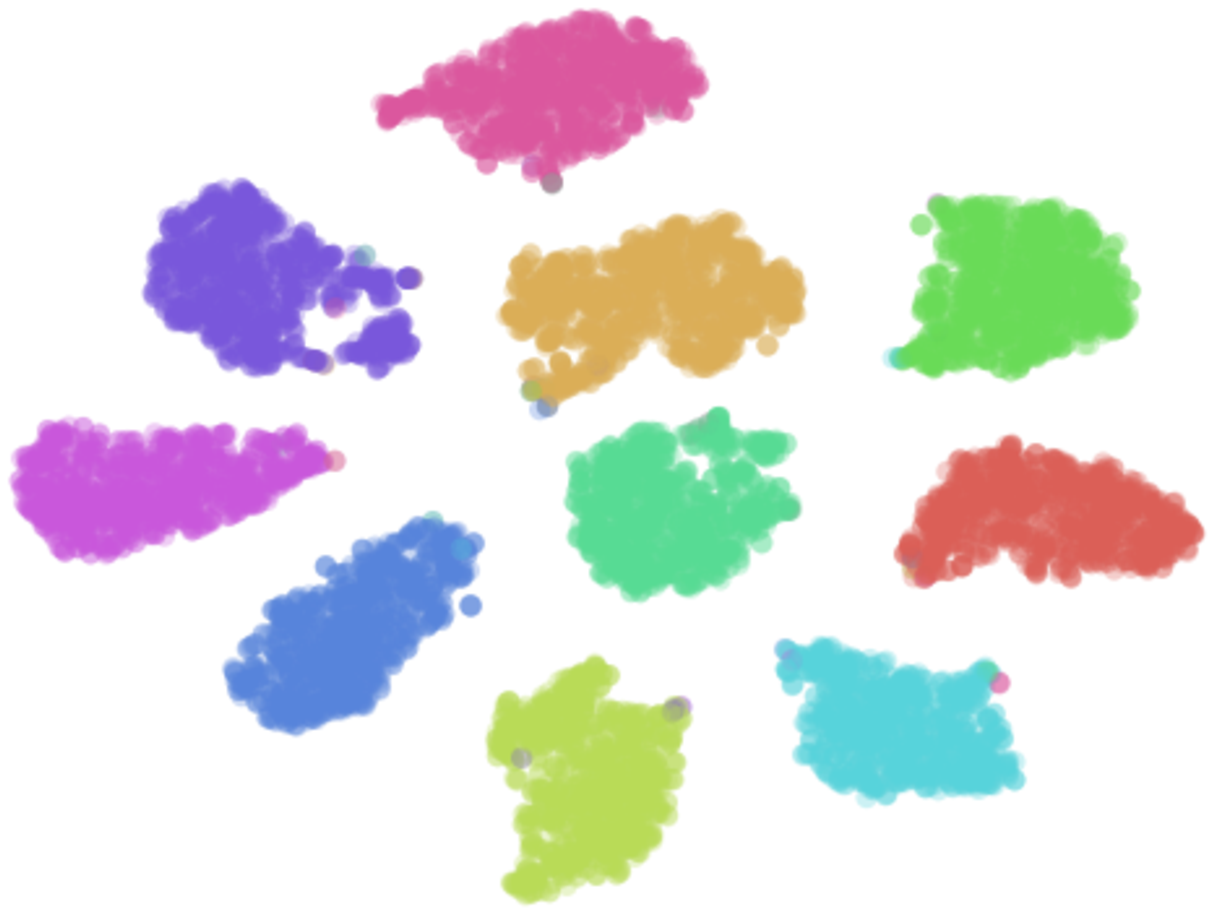}
    \includegraphics[width=0.32\textwidth]{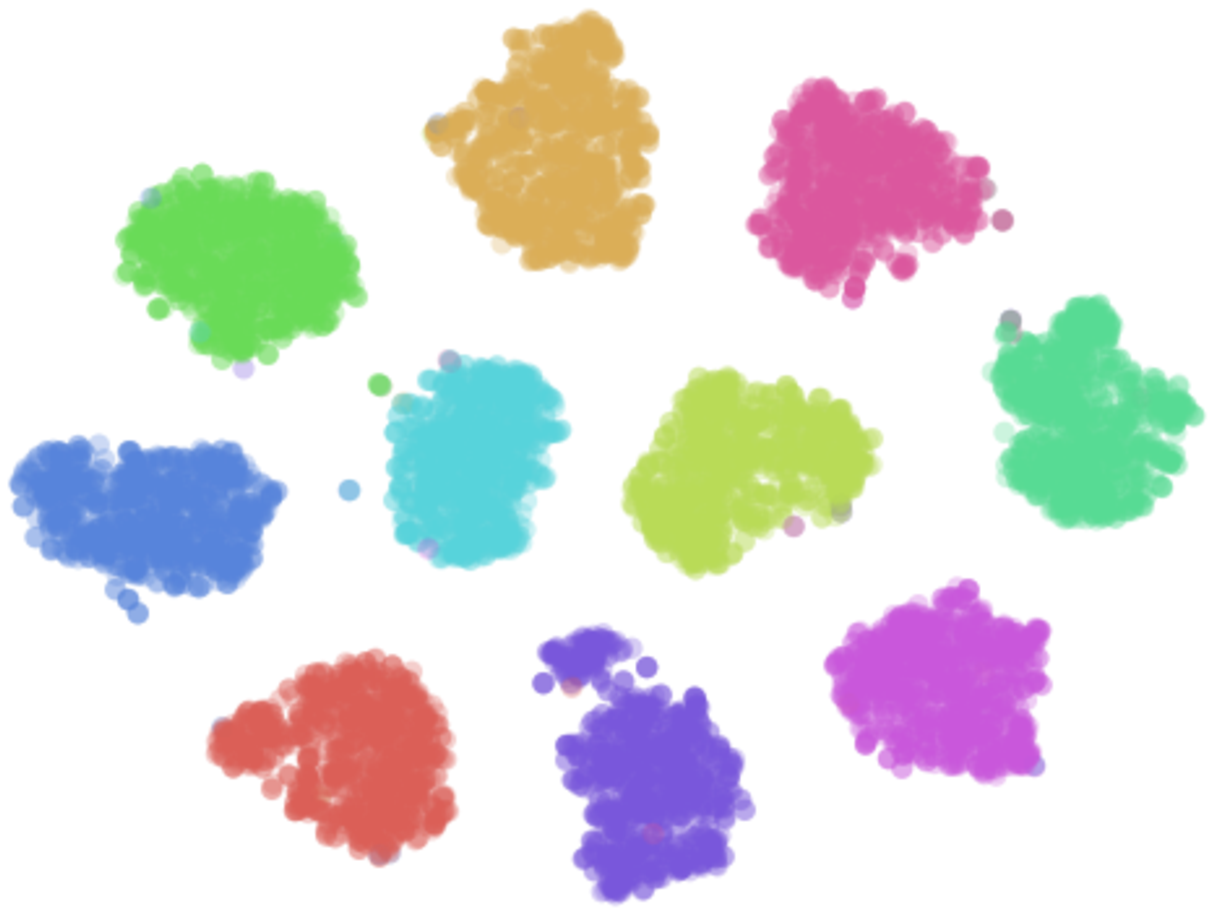}
  \caption{t-SNE visualization results on the features before the final FC layer. The input is the test set of MNIST. Left: trained with TisODE, middle: TisODE using a randomly chosen orthogonal matrix as the final FC, right: TisODE using proposed $\bV$ as the final FC.}
  \label{fig:tsne_q2}
\end{figure}

\begin{figure}[!htb]
\centering
    \includegraphics[width=0.32\textwidth]{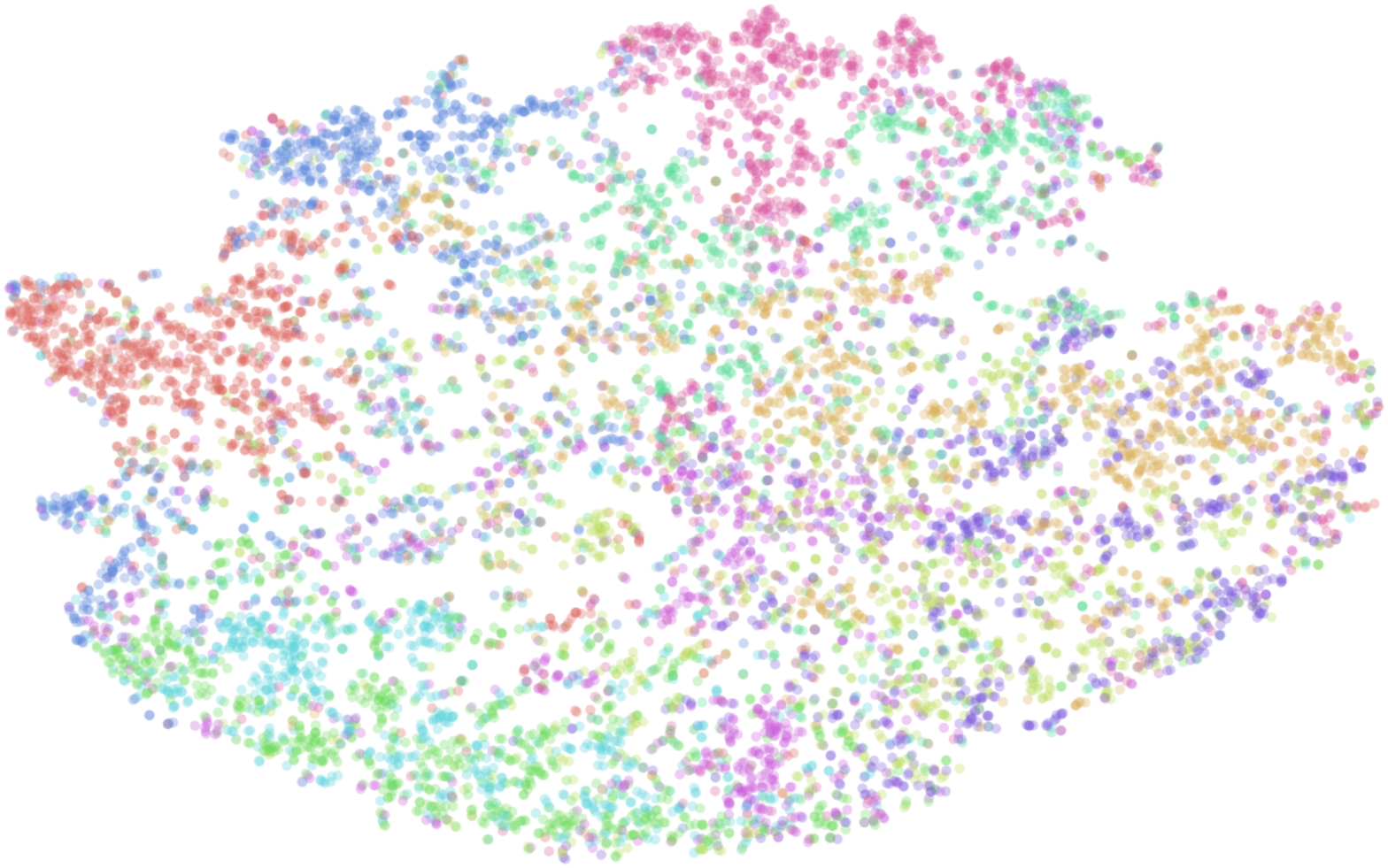}
    \includegraphics[width=0.32\textwidth]{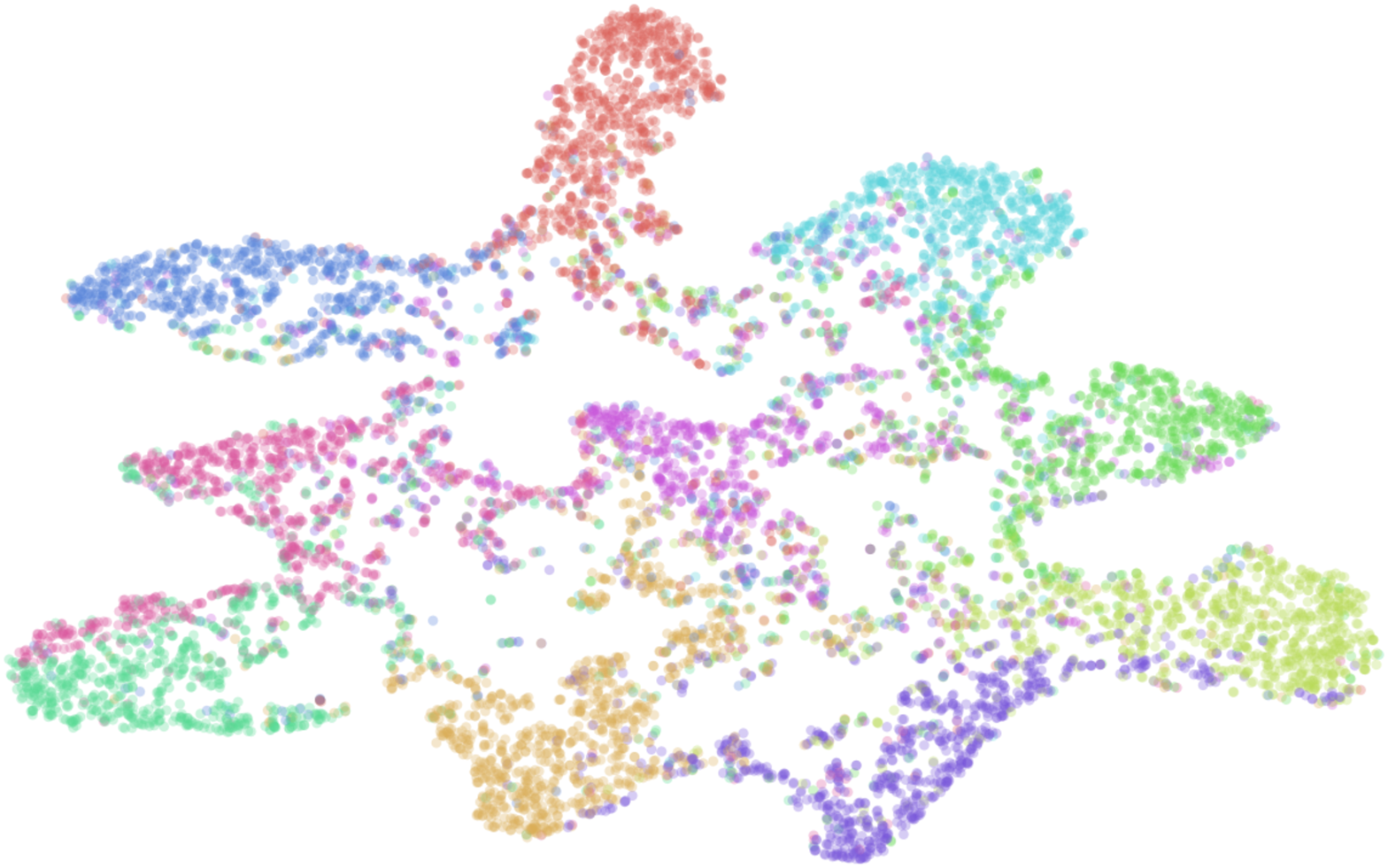}
    \includegraphics[width=0.32\textwidth]{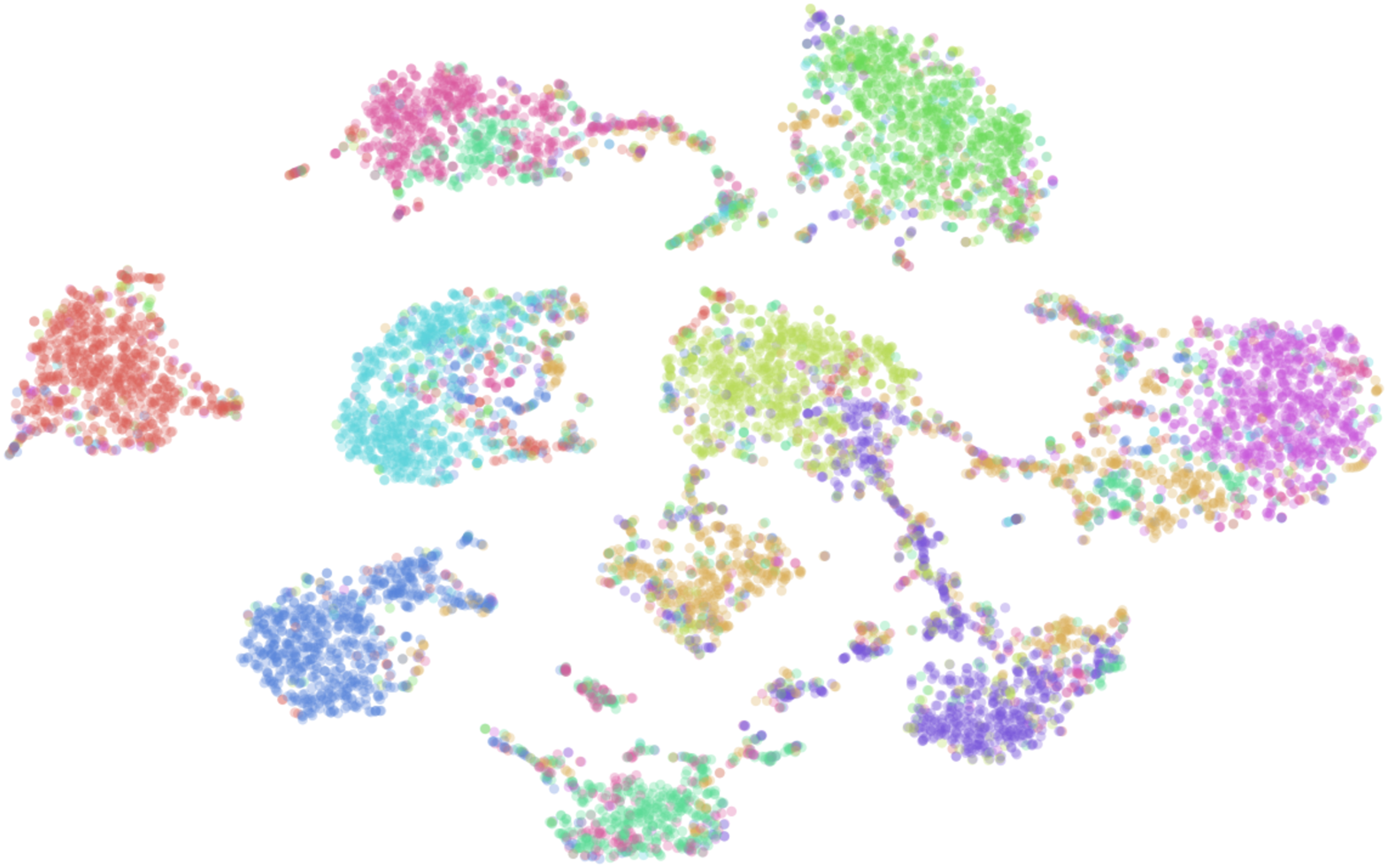}
  \caption{t-SNE visualization results on the features before the final FC layer. The input is the adversarial examples of the test set of MNIST generated using FGSM method at $\epsilon=0.3$. Left: trained with TisODE, middle: TisODE using a randomly chosen orthogonal matrix as the final FC, right: TisODE using proposed $\bV$ as the final FC.}
  \label{fig:tsne_q3}
\end{figure}

To validate our observations, we conduct experiments to compare the robustness of ODE net \cite{chen2018neural} and TisODE \cite{yan2019robustness} with and without our proposed FC layer $\bV$, on two standard datasets: MNIST \cite{LecunPIEEE1998} and CIFAR10 \cite{KriTR2009} \footnote{Our experiments are run on a GeForce RTX 2080 Ti GPU.}. On the MNIST dataset, all models consist of four convolutional layers and one fully-connected layer. On the CIFAR10 dataset, the networks are similar to those for MNIST except the down-sampling network is a stack of 2 ResNet blocks. In practice, the neural ODE can be
solved with different numerical solvers such as the Euler method and the Runge-Kutta methods \cite{chen2018neural}. Here, we use Runge-Kutta of order 5 in our experiments. Our implementation builds on the open-source neural ODE codes.\footnote{\url{https://github.com/rtqichen/torchdiffeq}} During training, no Gaussian noise or adversarial examples are augmented into the training set. 
{We test the performance of our model in defending against white-box attacks FGSM \cite{GoodICLR2015} and PGD \cite{MadryICLR2018} . The parameters for different attack methods used in this paper are given in the supplementary material.}
From \cref{tab:fc_1,tab:fc_2}, we observe that for both datasets, our fixed FC layer improves each network's defense ability by a significant margin. We visualize the features before the final FC layer using t-SNE \cite{van2008visualizing} in \cref{fig:tsne_q2,fig:tsne_q3}. We observe that with the FC layer, the features for different classes are well separated even under attacks.
{
\begin{table}[!thb]
\centering
\caption{Classification accuracy (\%) on adversarial MNIST examples, where the superscript $^+$ indicates the last FC layer is fixed to be $\bV$.} \label{tab:adv_mnist}
\vspace{-6pt}
\begin{tabular}{cccccc} 
\toprule
Attack & Para. & ODE & ODE$^+$  & TisODE & TisODE$^+$ \\
\midrule
None  & -             & 99.6 & 99.7  & 99.5 & 99.7   \\ 
\midrule                     
FGSM & $\epsilon=0.3$ & 31.4 & \textbf{52.8}  & 45.9 & \textbf{63.5}    \\ 
\midrule  
PGD  & $\epsilon=0.3$ & 0.29     & \textbf{0.30}   & 0.4  & \textbf{20.20}      \\ 
%\midrule  
%PGD  & $\epsilon=0.3$ &    &    &   &       \\ 
\bottomrule
\end{tabular}\label{tab:fc_1}
\end{table}

\begin{table}[!thb]
\centering
\caption{Classification accuracy (\%) on adversarial CIFAR10 examples, where the superscript $^+$ indicates the last FC layer is fixed to be $\bV$.} \label{tab:adv_cifar10}
\vspace{-6pt}
\begin{tabular}{cccccc} 
\toprule
Attack & Para. & ODE & ODE$^+$  & TisODE & TisODE$^+$ \\
\midrule
None  & -      & 87.0  & 85.0     & 87.4 & 81.8    \\ 
\midrule                     
FGSM & $\epsilon=0.1$ & 12.9    & \textbf{47.6}   & 13.1& \textbf{41.9}      \\ 
\midrule  
PGD  & $\epsilon=0.1$ & 7.8     & \textbf{14.7}   & 7.4  & \textbf{16.2}      \\ 
\bottomrule
\end{tabular}\label{tab:fc_2}
\end{table}
}

\subsection{Objective Formulation and Stability}
\label{subsect:SODEF_Formulation}

In this subsection, we formulate an optimization framework for SODEF to force output features to locate within the stable neighborhood of Lyapunov-stable equilibrium points. We make the following assumption.
\begin{Assumption}\label{assumpt:input}
The input $x$ takes values in a compact metric space $X$ and has probability distribution $\mu$. The feature extractor $h_{\bphi}$ is injective and continuous.
\end{Assumption}

The above assumption is satisfied if the input $x$ (e.g., an image) resides in a bounded and closed set of a Euclidean space. We denote the pushforward measure (still a probability distribution) of $\mu$ under the continuous feature extractor mapping $h_{\bphi}$ as $\nu_{\bphi} = \mu \circ h_{\bphi}^{-1}$, where $\circ$ denotes function composition. The conditional probability distribution for the embedding of each class $l\in\set{1,...,L}$ has compact support $E_l \subset\Real^n$ since $E_l$ is closed and $h_{\bphi}(X)$ is bounded in $\Real^n$. In \cref{subsect:max_dis}, the FC layer $\bV$ tries to maximize the distance between $E_l$, $l=1,\ldots,L$. In this section for analysis purposes, we also assume the following.

\begin{Assumption}\label{assumpt:disjoint}
We have $E_l\bigcap E_{l'} = \emptyset$ if $l\ne l'$, i.e., the supports of each class are pairwise disjoint.
\end{Assumption}

Our objective function is formulated as follows, which is explained in detail in the sequel:  
%\begin{align}
%\label{eq:obj}
%\min_{\btheta}&~ 
%\frac{1}{N}\sum_{i=0}^{N-1}  \ell(f_{\btheta}(z_i(T)),y_i))  \\
%{\rm s.t.} 
%&~ \frac{1}{N}\sum_{i=0}^{N-1} \norm{f_{\btheta}(z_i)}_{2}<\epsilon, f_{\btheta} \in C^{1}\nn
%&~ \frac{1}{N}\sum_{i=0}^{N-1} f'_{\btheta}(z_i)_{jj}<0 \quad \forall j =1,\dots, n \nn
%&~ \frac{1}{N}\sum_{i=0}^{N-1} (|f'_{\btheta}(z_i)|_{jj}-\sum_{k\ne j}|f'_{\btheta}(z_i)|_{jk})>0 \quad \forall j =1,\dots, n
%\end{align} %
\begin{align}
\label{eq:obj}
\min_{\btheta,\bphi}&~ 
\E_{\mu}\ell(\bV\T(\bz(T)), y_i) \\
\st 
&~ \E_{\nu_{\bphi}} \norm{f_{\btheta}(\bz(0))}_{2}<\epsilon,\ f_{\btheta} \in C^{1}(\Real^n,\Real^n),\label{eq:re1}\\
&~ \E_{\nu_{\bphi}}[\nabla f_{\btheta}(\bz(0))]_{ii} < 0,\ \forall i =1,\dots,n, \label{eq:re2}\\
&~ \E_{\nu_{\bphi}}\Big[\abs*{[\nabla f_{\btheta}(\bz(0))]_{ii}}-\sum_{j\ne i}\abs*{[\nabla f_{\btheta}(\bz(0))]_{ij}}\Big] >0,\ \forall i =1,\dots, n, \label{eq:re3}\\
&~ \bz(0)=h_{\bphi}(x), \text{ and } \bz(T)  \text{ is the output of \cref{eq:NODE} with input } \bz(0).  
\end{align}
Here, $\ell$ is a loss function and $\epsilon>0$ is a positive constant. The constraints \cref{eq:re1,eq:re2,eq:re3} force $\bz(0)$ to be near the Lyapunov-stable equilibrium points with strictly diagonally dominant derivatives. We limit the $f_{\btheta}$ to be in $C^1(\Real^n,\Real^n)$ to satisfy the condition in \cref{thm:linearization}. From \cite{hornik1991approximation}, we also know that standard multi-layer feed forward networks with as few as a single hidden layer and arbitrary bounded and non-constant activation function are universal approximators for $C^1(\Real^n,\Real^n)$ functions with respect to some performance criteria provided only that sufficiently many hidden units are available.%in $C^1$.
% From \cite{hornik1991approximation}, we know that with standard multi-layer feed forward networks with as few as a single hidden layer and arbitrary bounded and non-constant activation function are universal approximators for $C^m$ functions with respect to some performance criteria provided only that sufficiently many hidden units are available:

%\begin{Theorem}[Approximation Capabilities of Muitilayer Feedforward Networks Theorem\cite{hornik1991approximation}]
% For multi-layer feedforward networks with an arbitrarily large number of hidden units, if the activation function is non-constant and bounded and belongs to $C^m(\Real^n)$ (the space of all functions $f$ which, together with all their partial derivatives $D^{\alpha}f$ of order $\alpha\le m$ are continuous on $\Real^n$), then for all $f\in C^m(\Real^n)$, for all compact subsets $X$ of $R^n$, and for all $\epsilon > 0$, there is a network $f_{\btheta}$ with proper parameters such that $\norm{f-f_{\btheta}}_{m, X}< \epsilon$, where $\norm{f}_{m, X}\coloneqq \max_{\alpha\le m} \sup_{x\in X}|D^{\alpha}f|$.\label{thm:approximate}
% \end{Theorem}

As a comparison, TisODE \cite{yan2019robustness} only includes a constraint similar to $\cref{eq:re1}$, which in general provides no guarantee to force $\bz(0)$ near the Lyapunov-stable equilibrium points. In the extreme case with parameters $\btheta=0$ for $f_{\btheta}$ such that $f_{\btheta}=0$, the ODE degenerates to an identity mapping. No $\bz(0)\in\Real^n$ can now be a Lyapunov-stable equilibrium point, and no stability can therefore be guaranteed to defend against adversarial attacks even though the ODE curves still possess the non-intersecting property and steady-state constraint, which were cited as reasons for the stability of TisODE.

Instead of directly optimizing the above objective function, in our implementation, we optimize the following empirical Lagrangian with a training set $\{(x_k,y_k):k=1,...,N\}$:
\begin{align}\label{eq:obj_emp}
\min_{\btheta,\bphi}&\
\frac{1}{N}\sum_{k=0}^{N-1} \left( \ell\big(\bV\T\bz_k(T),y_k\big) + \alpha_1 \norm{f_{\btheta}\big(\bz_k(0)\big)}_{2} + \alpha_2 g_1\Big(\sum_{i=1}^{n} [\nabla f_{\btheta}(\bz_k(0))]_{ii}\Big)\right. \nn
& \quad+\left. \alpha_3 g_2 \Big(\sum_{i=1}^{n}(-\abs*{[\nabla f_{\btheta}(\bz_k(0))]_{ii}}+\sum_{j\ne i}\abs*{[\nabla f_{\btheta}(\bz_k(0))]_{ij}})\Big) \right)  \\
\st &\ \bz_k(0)=h_{\bphi}(x_k), \text{ and } \bz_k(T) \text{ is the output of \cref{eq:NODE} with input } \bz_k(0),\ \forall k=1,\ldots,N
\end{align}
where $\alpha_1$, $\alpha_2$ and $\alpha_3$ are hyperparameter weights, $g_1$ and $g_2$ are chosen monotonically increasing functions bounded below to eliminate the unbounded impact of the two regularizers that can otherwise dominate the loss. In this paper, we set $g_1(\cdot)=g_2(\cdot)=\text{exp}(\cdot)$. We call these two latter terms the SODEF regularizers.

%{we may also call the finitely many equilibrium stable points ``pitfall'' in each $E_L$.}
%{one question is whether \cref{lem:exist_f} and \cref{cor:discrete} is correct for countable values}

Suppose for each class $l=1,\ldots,L$, the embedding feature set $E_l=\set{\bz\tc{l}_1, \ldots, \bz\tc{l}_k}$ is finite. For each $i=1,\ldots,k$, let $\bA_i\in \Real^{n\times n}$ be strictly diagonally dominant matrix with every main diagonal entry be negative such that the eigenvalues for $\bA_i$ all have negative real part. From \cref{thm:Taussky}, each $\bA_i$ is non-singular and every eigenvalue of $\bA_i$ has negative real part. Therefore, from \cref{thm:stab_lin} and \cref{lem:exist_f}, there exists a function $f_{\btheta}$ such that all $\bz\tc{l}_i$ are Lyapunov-stable equilibrium points with corresponding first derivative $\nabla f_{\btheta}(\bz\tc{l}_i) = \bA_i$. This shows that if there exist only finite representation points for each class, we can find a function $f_{\btheta}$ such that all inputs to the neural ODE layer are Lyapunov-stable equilibrium points for $f_{\btheta}$ and 
\begin{enumerate}[(a)]
    \item\label[condition]{it:c1} $\E_{\nu_{\bphi}} \norm{f_{\btheta}(\bz(0))}_{2}=0$,
    \item\label[condition]{it:c2} $\E_{\nu_{\bphi}} [\nabla f_{\btheta}(\bz(0))]_{ii} < 0,\ \forall i =1,\dots,n$,
    \item\label[condition]{it:c3} $\E_{\nu_{\bphi}} [\abs*{[\nabla f_{\btheta}(\bz(0))]_{ii}} - \sum_{j\ne i} \abs*{[\nabla f_{\btheta}(\bz(0))]_{ij}} ] > 0,\ \forall i =1,\dots, n$.
\end{enumerate}

If the input space $X$ has infinite cardinality, then an injective and continuous feature extractor $h_{\bphi}$ results in a $\nu_{\bphi}$ with non-finite support, i.e., at least one $E_l$, $l=1,\ldots,L$, is infinite. It is not obvious whether we can obtain a $f_{\btheta}$ where every point in $E=\bigcup_l E_l$ is a stable equilibrium point. The following result gives a negative answer if $\nu_{\bphi}$ is a continuous measure (i.e., absolutely continuous \gls{wrt} Lebesgue measure) on some subset. 

\begin{Lemma}\label{lem:noexist_f}
If the restriction of $\nu_{\bphi}$ to some open set $E'\subset E$ is a continuous measure, there is no continuous function $f_{\btheta}$ such that for $\nu_{\bphi}$-almost surely all $\bz\in E$, $f_{\btheta}(\bz)=0$ and all the eigenvalues of $ \nabla f_{\btheta}(\bz)$ have negative real parts. In other words, there is no continuous function $f_{\btheta}$ such that almost surely all $\bz$ in $E$ are Lyapunov-stable equilibrium points.
\end{Lemma}

\cref{lem:noexist_f} indicates that it is too much to hope for all points in $E$ to be Lyapunov-stable equilibrium points. In the following, we relax this requirement and show that under mild conditions, for all $\epsilon>0$, we can find a continuous function $f_{\btheta}$ with finitely many stable equilibrium points such that \cref{it:c2,it:c3} above hold and \cref{it:c1} is replaced by $\E_{\nu_{\bphi}} \norm{f_{\btheta}(\bz(0))}_{2}<\epsilon$. This motivates the optimization constraints in \cref{eq:re1,eq:re2,eq:re3}.

\begin{Theorem}\label{thm:exist_f} 
Suppose \cref{assumpt:input,assumpt:disjoint}. If $\nu_{\bphi}$ is not a continuous uniform measure on $E_l$ for each $l=1,\ldots,L$, then the following holds: 1) The function space satisfying the constraints in \cref{eq:re1,eq:re2,eq:re3} is non-empty for all $\epsilon>0$. 2) If additionally the restriction of $\nu_{\bphi}$ to any open set $O \subset E_l$ is not a continuous uniform measure, there exist functions in this space such that each support $E_l$ contains at least one Lyapunov-stable equilibrium point.
\end{Theorem}

% {[not sure if we can prove \cref{thm:high_Pr}, but have some ideas and is trying, please ignore this theorem currently]
% Furthermore, if $\nu_{\bphi | l}$ is absolutely continuous \gls{wrt} Lebesgue measure but it is not a uniform distribution, in the following theorem we know that under mild assumptions there exist a $f$ such that $\bz(0)$ is in a stable neighborhood of Lyapunov-Stable equilibrium points with arbitrary high probability.
% \begin{Theorem}\label{thm:high_Pr}
% %We assume $f_{\btheta}$ is a $C^{2}$ mapping, 
% For any $\delta>0$, we can find $f$ that satisfies the constraints in \cref{eq:re1,eq:re2,eq:re3} such that $\bz(0)$ is within a stable neighborhood of Lyapunov-Stable Equilibrium Points with probability larger than 1-$\delta$ \gls{wrt} $\nu_{\bphi}$.
% \end{Theorem}
% }

\section{Experiments}
\label{sect:exper}

In this section, we evaluate the robustness of SODEF under adversarial attacks with different attack parameters. We conduct experiments to compare the robustness of ODE net \cite{chen2018neural} and TisODE net \cite{yan2019robustness} on three standard datasets: MNIST \cite{LecunPIEEE1998}, CIFAR10 and CIFAR100 \cite{HintonSPM2012}. Since SODEF is compatible with many defense methods, it can be applied to any neural network's final regressor layer to enhance its stability against adversarial attacks. {Our experiment codes are provided in \url{https://github.com/KANGQIYU/SODEF}.}

%{Additional experimentally results like combining adversarial training and SODEF and ablation study are provided in the supplementary material.}

\subsection{Setup}

We use open-source pre-trained models that achieve the top accuracy on each dataset as the feature extractor $h_{\bphi}$. Specifically for simple MNIST task, we use the ResNet18 model provided in Pytorch. We use the model provided by \cite{Luo_DCL_2019}, which obtains nearly $88\%$ clean accuracy on CIFAR100 using EfficientNet \cite{tan2019efficientnet} and the model provided by \cite{pytorchcifar10}, which has nearly $95\%$ clean accuracy on CIFAR10. In the neural ODE layer, $f_{\btheta}$ consists of 2 FC layers. During the trainings of SODEF (except in the experiment included in \cref{subsec:com_trades}), we train the neural network with the fixed FC introduced in \cref{subsect:max_dis}. In the first $30$ epochs, we fixed $f_{\btheta}$ to let the feature extractor $h_{\bphi}$ learn a feature representation with only the cross-entropy loss $\ell$, and in the remaining $120$ epochs, we release $h_{\bphi}$ to further train $f_{\btheta}$ using \cref{eq:obj_emp} with $\alpha_1=1$ and $\alpha_2=\alpha_3=0.05$. {For CIFAR10 and CIFAR100, the pixel values are normalized by $(x-\mu)/\sigma$ where $\mu=[0.4914, 0.4822, 0.4465]$ and $\sigma=[0.2023, 0.1994, 0.2010]$ \footnote{To test AutoAttack, we have strictly followed the instruction in \url{https://github.com/RobustBench/robustbench} to attack the original images before any normalization or resizing.}. To show that our SODEF is compatible with many defense methods and can be applied to any neural network's final regression layer, we conduct an experiment where we use a recently proposed robust network TRADES \cite{pang2020bag} as the feature extractor in our SODEF. The pretrained model is provided here \footnote{https://github.com/P2333/Bag-of-Tricks-for-AT}, and we choose the model with architecture "WRN-34-10" to conduct our experiments. Besides the two vanilla white-box attacks FGSM and PGD as metioned in \cref{subsect:max_dis}, we also include a strong ensemble attack AutoAttack \cite{croce2020reliable}, which sequentially performs attack using all of the following four individual attacks: three white-box attacks APGD$_{\text{CE}}$,  APGD$^{\text{T}}_{\text{DLR}}$ and FAB$^{\text{T}}$\cite{croce2020minimally}, and one black-box  Square attack \cite{andriushchenko2020square}. We refer the reader to the the supplementary material for more details of the attacks used in this paper, where, in additional, more experiments are included.} 

%In our following experiments, we set the feature extractor to be the open-source model mentioned above with {the neural ODE layer} being ODE net, TisODE or SODEF.

{\subsection{Compatibility of SODEF} \label{subsec:com_trades}
 Adversarial training (AT) is one of the most effective strategies for defending adversarial attacks. 
 TRADES \cite{pang2020bag} is one of the adversarial training defense methods with combinations of tricks of warmup, early stopping, weight decay, batch size and other  hyper parameter settings. In this experiment we fix the pretained TRADES model (except the final FC layer (size 640x10)) as our feature extractor $h_{ \phi }$. We then append our (trainable) SODEF with integration time $T=5$ to the output of the feature extractor. To evaluate model robustness, we use AutoAttack and attack the models using both the $\mathcal{L}_2$ norm ($\epsilon=0.5$) and   $\mathcal{L}_{\infty}$ norm ($\epsilon=8/255$). The results are shown in \cref{tab:r2_3}. We clearly observe that our SODEF can enhance TRADES's robustness under all the four individual attacks and the strongest ensemble AutoAttack. For the strong $\calL_2$ AutoAttack, our SODEF have improved the model robustness from $59.42\%$ to $67.75\%$. Our experiment show that SODEF can be applied to many defense models' regression layer to enhance their stability against attacks.
\begin{table}[!tbh]
\centering
\caption{Classification accuracy (\%) using TRADES (w/ and w/o SODEF) under AutoAttack  on adversarial CIFAR10 examples with  $\mathcal{L}_2$ norm ($\epsilon=0.5$) and   $\mathcal{L}_{\infty}$ norm ($\epsilon=8/255$).} 
\vspace{-6pt}\label{tab:r2_3}
\small
\setlength{\tabcolsep}{0pt}
\begin{tabular}{ccccccc} 
\toprule
Attack / Model       &TRADES $\mathcal{L}_{\infty}$  &TRADES+SODEF $\mathcal{L}_{\infty}$ & TRADES $\mathcal{L}_2$ & TRADES+SODEF $\mathcal{L}_2$ \\
\midrule
%\midrule 
\multirow{1}{*}{Clean}                           &  85.48 &  85.18  &85.48 &85.18 \\   
\multirow{1}{*}{APGD$_{\text{CE}}$}              & 56.08  &\tb{ 70.90}   & 61.74      & \tb{74.35}\\   
\multirow{1}{*}{APGD$^{\text{T}}_{\text{DLR}}$}  & 53.70  & \tb{64.15}     & 59.22     &\tb{68.55} \\  
\multirow{1}{*}{FAB$^{\text{T}}$}                & 54.18  & \tb{82.92}     & 60.31     &\tb{83.15}\\  
\multirow{1}{*}{Square}                           & 59.12  &\tb{62.21}     & 72.65     & \tb{76.02} \\ 
\multirow{1}{*}{AutoAttack}                      & 53.69  & \tb{57.76}  & 59.42    & \tb{67.75}\\
\bottomrule
\end{tabular}
\end{table}
\subsection{Influence of Integration Time $T$}
From the discussion after \cref{thm:linearization,thm:stab_lin},  we know if the malicious perturbations around the ODE input Lyapunov-stable equilibrium point $\bz(0)$ is small, then the output  $\bz(T)$ for large enough $T$ will not be affected significantly by the perturbation: $ \norm{\tilde{\bz}(t)-\bz(0)}_2\to 0 \text{ as } t\to \infty$. Consequently, the succeeding network layers after the neural ODE layer can still perform well without being affected by the input perturbation. In this section, we test the influence of the SODEF integration time $T$ using CIFAR100. We use the model EfficientNet provided by   \cite{tan2019efficientnet} as $h_{ \phi }$ (Note, unlike \cref{subsec:com_trades}, $h_{ \phi }$ is trainable in this experiments). We use  AutoAttack with  $\mathcal{L}_{2}$ norm ($\epsilon=0.5$). We observe that for all the four individual attacks and the strongest ensemble AutoAttack, SODEF  performs generally better for large integration time $T$. We also test larger integration time $T>10$, but do not see any obvious improvements.
\begin{table}[!tbh]
\centering
\caption{Classification accuracy (\%) under AutoAttack  on adversarial CIFAR100 examples with  $\mathcal{L}_2$ norm, $\epsilon=0.5$ and different integration time $T$ for SODEF.} \label{tab:r3_2}
\vspace{-6pt}
\small
\begin{tabular}{ccccccccccc} 
\toprule
Attack / $T$       &1  &3  & 5   & 6 &7  &8  &9  & 10 \\
\midrule
%\midrule 
\multirow{1}{*}{Clean}                          &  88.00 & 88.12      &88.15          &   88.00         &  87.92  &88.00  &88.05 &88.10  \\   
\multirow{1}{*}{APGD$_{\text{CE}}$}              &  17.20 & 21.33      &21.05          &   23.67      &  69.67  &85.33  &\tb{87.10} &86.88  \\   
\multirow{1}{*}{APGD$^{\text{T}}_{\text{DLR}}$}  &  21.02 & 21.00     &22.00            &  26.00       &  63.30   &\tb{86.90}  &{86.20} &86.54    \\   
\multirow{1}{*}{FAB$^{\text{T}}$}                &  86.33 & 85.10      &86.36            &  \tb{87.70}      &  87.67  &86.55  &86.22 &85.93    \\   
\multirow{1}{*}{Square}                           &  84.67 & 86.22       & 87.05          &  87.20       &  86.90    &86.33  &\tb{87.05} &86.75  \\   
\multirow{1}{*}{AutoAttack}                      &  2.00 & 3.53      &4.87           &  4.33      &  30.66   &78.80  &78.97 &\tb{79.10}     \\   
\bottomrule
\end{tabular}
\end{table}

\subsection{Performance Comparison Under AutoAttack}
For a comparison, we provide the results of applying AutoAttack to other baseline models mentioned in the paper. We set the same integration time  for ODE, TisODE and SODEF. We observe that for the strongest AutoAttack, our SODEF outperforms the other baseline models  by a significant margin. In this case, SODEF achieves $79.10\%$ accuracy while other models only get less than $3\%$ accuracy.

\begin{table}[!tbh]
\centering
\caption{Classification accuracy (\%) under AutoAttack  on adversarial CIFAR100 examples with  $\mathcal{L}_2$ norm, $\epsilon=0.5$ and $T=10$.} 
\vspace{-6pt}\label{tab:r3_3}
\small
\begin{tabular}{ccccccc} 
\toprule
Attack / Model       &No ODE &ODE & TisODE & SODEF\\
\midrule
%\midrule 
\multirow{1}{*}{Clean}                           &  88.00 &  87.90  &88.00 &88.10 \\   
\multirow{1}{*}{APGD$_{\text{CE}}$}              & 23.30  & 6.75   & 14.32    &\tb{86.88} \\   
\multirow{1}{*}{APGD$^{\text{T}}_{\text{DLR}}$}  & 7.33  & 22.00   & 24.20    &\tb{86.54}\\  
\multirow{1}{*}{FAB$^{\text{T}}$}                & 79.30  & 78.67   & 77.16    &\tb{85.93}\\  
\multirow{1}{*}{Square}                           & 84.52  &85.67  & 86.32    &\tb{86.75 } \\ 
\multirow{1}{*}{AutoAttack}                      & 0.00  & 1.33   & 4.06    &\tb{79.10} \\
\bottomrule
\end{tabular}
\end{table}

}
\subsection{Performance Under PGD and FGSM Attacks}
White-box adversaries have knowledge of the classifier models, including training data, model architectures and parameters. We test the performance of our model in defending against the white-box attacks, PGD and FGSM. We set $T=5$ as the integration time for the neural ODE layer. The parameters for different attack methods used are given in the supplementary material. The subsequent experiments use these settings by default, unless otherwise stated.

% \begin{figure}[!htb]
% \centering
%     \includegraphics[width=0.32\textwidth]{jacobian_ODE.png}
%     \includegraphics[width=0.32\textwidth]{jacobian_SODEF.png}
%     \includegraphics[width=0.32\textwidth]{denseattack.png}

%   \caption{Eigenvalues visualization for the Jacobian matrix $\evalat*{[\nabla f_{\btheta}]}{\bz(0)}$ from different ODEs. Left: trained with SODEF,  right: trained with ODE net.}
%   \label{fig:ecn_q2}
% \end{figure}

% \begin{figure}[!htb]
% \centering
%   \caption{Eigenvalues visualization for the Jacobian matrix $\evalat*{[\nabla f_{\btheta}]}{\bz(0)}$ from different ODEs. Left: trained with SODEF,  right: trained with ODE net.}
%   \label{fig:ecn_q2}
% \end{figure}

\begin{table}[!tbh]
\centering
\caption{Classification accuracy (\%) on adversarial MNIST examples.} \label{tab:adv_mnist_SOEEF}
\vspace{-6pt} 
\small
\begin{tabular}{cccccccc} 
\toprule
Attack & Para.                          &no ode   &  ODE &   TisODE   & SODEF \\
\midrule
None  & -                               &99.45          &99.42        &     99.43              &    99.44        \\ 
%\midrule       
\multirow{1}{*}{FGSM}  &  $\epsilon=0.3$ & 10.03         & 29.6             &   {36.70}          &    {\bf 63.36}      \\   
\multirow{1}{*}{PGD}  &  $\epsilon=0.3$ & 0.31           &1.56          &        {1.82}          &    {\bf 45.25}      \\   
%\midrule 
%\multirow{1}{*}{C\&W} &  $\kappa=1$    &  ?                  &        ?          &   {\bf 42.1}    \\ 
%\midrule    
%\multirow{1}{*}{BSA} & $\alpha=0.8$      & ?                  &     ?             & ?  \\ 
%\midrule
%\multirow{1}{*}{JSMA} & $\gamma=0.6$      & ?                &     ?             &?  \\ 
%\midrule
%\# params & -              &  818,334              &  401,168          & 490,209   \\
\bottomrule
\end{tabular}
\end{table}

The classification results on MNIST are shown in \cref{tab:adv_mnist_SOEEF}. We observe that while maintaining the state-of-the-art accuracy on normal images, SODEF improves the adversarial robustness as compared to the other two methods. For the most effective attack in this experiment, i.e., PGD attack, SODEF shows a $45.25\%-1.56\%=43.69\%$ improvement over ODE and a $45.25\%-1.23\%=44.02\%$ improvement over TisODE.

\begin{table}[!tbh]
\centering
\caption{Classification accuracy (\%) on adversarial CIFAR10 examples.} \label{tab:adv_cifar10_SODEF}
\vspace{-6pt}
\small
\begin{tabular}{cccccccc} 
\toprule
Attack & Para.                          &no ode   & ODE  & TisODE  & SODEF \\
\midrule
None  & -                               & 95.2                 &     94.9         & 95.1    &    95.0        \\ 
%\midrule       
\multirow{1}{*}{FGSM}  &  $\epsilon=0.1$ & 47.31                &        45.23    &  43.28     &    {\bf 68.05}      \\   
\multirow{1}{*}{PGD}  &  $\epsilon=0.1$ & 3.09                   &        3.21     &3.80      &    {\bf 55.59}      \\   
%\midrule 
%\multirow{1}{*}{C\&W} &  $\kappa=1$    &  ?                  &        ?          &   {\bf ?}    \\ 
%\midrule    
%\multirow{1}{*}{BSA} & $\alpha=0.8$      & ?                  &     ?             & ?  \\ 
%\midrule
%\multirow{1}{*}{JSMA} & $\gamma=0.6$      & ?                &     ?             &?  \\ 
%\midrule
%\# params & -              &  818,334              &  401,168          & 490,209   \\
\bottomrule
\end{tabular}
\end{table}

For CIFAR-10, we see from \cref{tab:adv_cifar10_SODEF} that SODEF  maintains high accuracy on normal examples and makes the best predictions under adversarial attacks. In particular, SODEF achieves an absolute percentage point improvement over ODE net up to $52.38\%$ and over TisODE up to $52.54\%$ for PGD attack. 

For CIFAR-100, %\cref{tab:adv_cifar100_SODEF} 
the results in the supplementary material
shows that the most effective attack causes the classification accuracy to drop relatively by $74.6\%=\frac{88.0-22.35}{88.0}$  for SODEF and by $97.3\%=\frac{88.3-2.39}{88.3}$ for vanilla EfficientNet, which is pre-trained on ImageNet to obtain a top clean accuracy. Neither ODE net nor TisODE net can improve the classification accuracy under PGD attack by a big margin, e.g. TisODE net only improves the classification accuracy from $2.39\%$ to $3.44\%$, while SODEF still shows clear defense capability in this scenario.

\subsection{Ablation Studies}

The impact of the ODE with and without the SODEF regularizers in \cref{eq:obj_emp} has been presented in the above comparisons between SODEF and ODE. In this section, we show the necessity of diversity promoting using the FC introduced in \cref{subsect:max_dis} and conduct transferability study.

\subsubsection{Impact of Diversity Promotion}

\begin{table}[!htb]
\centering
\caption{Classification accuracy (\%) on adversarial MNIST examples, where the superscript $^-$ indicates the last FC layer is not fixed to be $\bV$ and is set to be a trainable layer. } \label{tab:adv_mnist_fixfc_ornot}
\begin{tabular}{cccc} 
\toprule
Attack & Para. & SODEF & SODEF$^-$ \\
\midrule
None  & -             & 95.0 & 95.1    \\ 
\midrule                     
FGSM & $\epsilon=0.1$ & {\bf 63.36} & 51.6       \\ 
\midrule  
PGD  & $\epsilon=0.1$ & {\bf 45.25}  & 34.9     \\ 
\bottomrule
\end{tabular}
\end{table}

\cref {tab:adv_mnist_fixfc_ornot} shows the difference of the defense performance when fixing the final FC be $\bV$ or setting it to a trainable linear layer. It can be seen that having diversity control improves the robustness. One possible reason for this phenomenon given in \cref{sect:SODEF} is that diversity promotion with a fixed designed FC attempts to make the embedding feature support $E_l$ of each class $l$ disjoint to each other and therefore the Lyapunov-stable equilibrium points for each $E_l$ are well separated.

\subsubsection{Transferability Study}
Transferability study is carried out on CIFAR-10, where the adversarial examples are generated using FGSM and PGD attacks using ResNet18 without any ODEs. The classification accuracy drops from $68.05\%$ to $59\%$ for FGSM with $\epsilon=0.3$, and from $55.59\%$ to $34\%$ for PGD with $\epsilon=0.1$. One possible reason for this phenomenon is that ODEs have obfuscated gradient masking effect as discussed in \cite{huang2020adversarial}, and a transfer attack may deteriorate the defense effect. However, as we observe from \cref{tab:adv_cifar10_SODEF}, even with a transfer attack on SODEF, it still performs better than other ODEs without transfer attacks.

\section{Conclusion}\label{sect:conc}
In this paper, we have developed a new neural ODE network, SODEF, to suppress input perturbations. SODEF is compatible with any existing neural networks and can thus be appended to the state-of-the-art networks to increase their robustness to adversarial attacks. We demonstrated empirically and theoretically that the robustness of SODEF mainly derives from its stability and proposed a training method that imposes constraints to ensure all eigenvalues of the Jacobian matrix of the neural ODE layer have negative real parts. When each classification class converges to its own equilibrium points, we showed that the last FC layer can be designed in such a way that the distance between the stable equilibrium points is maximized, which further improves the network's robustness. The effectiveness of SODEF has been verified under several popular while-box attacks.
\begin{ack}
This research is supported in part by A*STAR under its RIE2020 Advanced Manufacturing and Engineering (AME) Industry Alignment Fund – Pre Positioning (IAF-PP) (Grant No. A19D6a0053) and the RIE2020 Industry Alignment Fund – Industry Collaboration Projects (IAF-ICP) Funding Initiative, as well as cash and in-kind contribution from the industry partner(s). The computational work for this article was partially performed on resources of the National Supercomputing Centre, Singapore (https://www.nscc.sg).
\end{ack}
\section*{Broader Impact}
Our work, which contributes to more robust DNNs, is supposed to mitigate the threat of adversarial attacks. However, on the hand, the reliable deployment of DNNs in automation of tasks will potentially bring mass-scale unemployment and social unrest. As DNNs become more robust and more tasks, especially those whose failures will bring high risks to human lives or large property losses under adversarial attacks, fall into the automatic task category, massive jobs could disappear.

\appendix

\section{Related Work and Adversarial Attacks}
In this section, we give an overview of related work in stable neural ODE networks. We also give an overview of common adversarial attacks and recent works that defend against adversarial examples. 
 
\subsection*{Stable Neural Network}
Gradient vanishing and gradient exploding are two well-known phenomena in deep learning \cite{Ben1994gd}. The gradient of the objective function, which strongly relies on the training method as well as the neural network architecture, indicates how sensitive the output is \gls{wrt} input perturbation. Exploding gradient implies instability of the output \gls{wrt} the input and thus resulting in a non-robust learning architecture. On the other hand, vanishing gradient implies insensitivity of the output \gls{wrt} the input, i.e., robustness against input perturbation. However, this prohibits effective training. To overcome these issues, \cite{Hab2018stable,huang2020adversarial} proposed a new DNN architecture inspired by ODE systems. This DNN architecture is stable in the sense that the input-output of the linearized system is always norm-preserving. This is different from the Lyapunov stability \cite{chen1999linear} our paper pursues. 

The reference \cite{Li2020IEskip} developed an implicit Euler based skip-connection to enhance ResNets with better stability and adversarial robustness. More recently, \cite{yan2019robustness} explored the robustness of neural ODE and related the robustness of neural ODE against input perturbation with the non-intersecting property of ODE integral curves. In addition, \cite{yan2019robustness} proposed TisODE, which regularizes the flow on perturbed data via the time-invariant property and the imposition of a steady-state constraint. However, in general, neither the non-intersecting property nor the steady-state constraint can guarantee robustness against input perturbation. In this paper, we  propose to impose constraints to ensure Lyapunov stability during training so that adversarial robustness can be enhanced. 

\subsection*{Adversarial Examples}
Recent works \cite{SzegedyICLR2013,sharif2016accessorize,moosavi2016deepfool} demonstrated that in image classification tasks, the input images can be modified by an adversary with human-imperceptible perturbations so that a DNN is fooled into mis-classifying them. In \cite{carlini2018audio,schonherr2018adversarial}, the authors proposed targeted attacks for audio waveform tasks. They showed that adversarial audio can be embedded into speech so that DNNs cannot recognize the input as human speech. Furthermore, adversarial speech can fool speech-to-text DNN systems into transcribing the input into any pre-chosen phrase. The references \cite{xiang2019generating,cao2019adversarial} proposed to generate targeted unnoticeable adversarial examples for 3D point clouds, which threatens many safety-critical applications such as autonomous driving.

Adversarial attacks \cite{GoodICLR2015,KurakinICLR2017a,MadryICLR2018,PapernotESSP,CarliniISSP2017} can be grouped into two categories: white-box attacks where adversaries have knowledge of the classifier models, including training data, model architectures and parameters, and black-box attacks where adversaries do not know the model's internal architecture or training parameters. An adversary crafts adversarial examples based on a substitute model and then feed these examples to the original model to perform the attack. 

To mitigate the effect of adversarial attacks, many defense approaches have been proposed such as adversarial training \cite{SzegedyICLR2013, GoodICLR2015},  defensive distillation \cite{PapISSP2016}, ECOC based methods \cite{VermaNIPS2019,SongAAAI2021}, and post-training defenses \cite{HendrycksICLR2017,MengCCS2017,SamICLR2018}. None of these methods make use of the stability of ODEs. 

For a normal image-label pair $(x,y)$ and a trained DNN $f_{\btheta}$ with $\btheta$ being the vector of trainable model parameters, an adversarial attack attempts to find an adversarial example $x'$ that remains within the $\calL_p$-ball with radius $\epsilon$ centered at the normal example $x$, i.e., $\|x-x'\|_p\leq \epsilon$, such that $f_{\btheta}(x') \neq y$. In what follows, we briefly present some popular adversarial attacks that are used to verify the robustness of our proposed SODEF.

\textit{Fast Gradient Sign Method (FGSM)} \cite{GoodICLR2015} perturbs a normal input $x$ in its $\calL_\infty$ neighborhood to obtain 
\begin{align*}
x'=x+\epsilon \cdot {\rm sign}\left(\nabla_x \calL(f_{\btheta}(x),y)\right),
\end{align*} 
where $\calL(f_{\btheta}(x),y)$ is the cross-entropy loss of classifying $x$ as label $y$, $\epsilon$ is the perturbation magnitude, and the update direction at each pixel is determined by the sign of the gradient evaluated at this pixel.
FGSM is a simple yet fast and powerful attack. 

\textit{Projected Gradient Descent (PGD)} \cite{MadryICLR2018}  iteratively refines FGSM by taking multiple smaller steps $\alpha$ in the direction of the gradient. The refinement at iteration $i$ takes the following form:
\begin{align*}
x'_i=x'_{i-1} + {\rm clip}_{\epsilon,x}\left(\alpha \cdot {\rm sign}\left(\nabla_x \calL(f_{\btheta}(x),y)\right)\right),
\end{align*}
where $x'_0=x$ and ${\rm clip}_{\epsilon,x}(x')$ performs clipping of the input $x'$. For example, in an image $x'$, if a pixel takes values between 0 and 255, ${\rm clip}_{\epsilon,x}(x')=\min\left\{255,x+\epsilon,\max\{0,x-\epsilon,x'\}\right\}$.

{\textit{AutoAttack:} In
\cite{croce2020reliable}, the authors
proposed two extensions of the PGD-attack, named APGD$_{\text{CE}}$ and APGD$^{\text{T}}_{\text{DLR}}$, to overcome failures of vanilla PGD due to suboptimal step size and
problems of the objective function. They combined these two attacks with two complementary
attacks called (white-box) FAB$^{\text{T}}$\cite{croce2020minimally} and (black-box) Square\cite{andriushchenko2020square} to form a parameter-free ensemble of attacks, called AutoAttack, to test adversarial robustness.}
% \textbf{ Carlini \& Wagner (C\&W)} \cite{CarliniISSP2017}:  
% In this paper, we consider the C\&W $L_2$ attack only. Let $x'=x+\frac{1}{2}(\tanh(\omega)+1)$, where the operations are performed pixel-wise. For a normal example $(x,y)$, C\&W  finds the attack perturbation by solving the following optimization problem
% \begin{align*}
% \underset{\omega}{\min} \ \left\| \frac{1}{2}(\tanh(\omega)+1)-x \right\|^2_2+c\cdot \ell\left(\frac{1}{2}(\tanh(\omega)+1)\right),
% \end{align*}
% where $\ell(x')=\max\left(Z(x',y)-\max\{Z(x',i):i\neq y\}+\kappa, 0\right)$. A large confidence parameter $\kappa$ encourages misclassification.

%Furthermore with arbitrarily large number of hidden units there exist a continuous function $f_{\btheta}$ such that the above conclusions are correct.
\section{Attack Parameters}

For two vanilla white-box {FGSM and PGD} attacks performed in \cref{subsect:max_dis,sect:exper}, {we use $\mathcal{L}_\infty$ norm with maximum perturbation $\epsilon=0.3$ for MNIST and $\epsilon=0.1$ for CIFAR10 and we iterates PGD for 20 times with step size 0.1.} {For the ensemble AutoAttack, we use the standard $\mathcal{L}_2$ norm with maximum perturbation $\epsilon=0.5$ and   $\mathcal{L}_{\infty}$ norm with maximum perturbation $\epsilon=8/255$ as in the benchmark\footnote{https://robustbench.github.io/}}

{
\section{Influence of Attack Strength}
We note that in adversarial attacks, perturbations are assumed to be imperceptible \cite{SzegedyICLR2013}, otherwise attack detection techniques can be used \cite{MengCCS2017}. Therefore, most related literature on robust techniques against adversarial attacks assume small perturbation. However, it is hard to obtain a theoretical analysis for an exact perturbation bound, which highly depends on specific datasets and model parameters. We instead demonstrate it experimentally.  To show how much perturbation over input $x$ the neural network can defend against, we conduct an experiment here using SODEF where we increase the FGSM attack strength $\epsilon$ at input $x$. 
%All parameters used here, except $\epsilon$, are the same as in \cref{tab:adv_para} in the supplementary material. 
Note that in the paper, for CIFAR10 and CIFAR100, the pixel values are normalized by $(x-\mu)/\sigma$ where $\mu=[0.4914, 0.4822, 0.4465]$ and $\sigma=[0.2023, 0.1994, 0.2010]$.  The attack results are shown in \cref{fig:diff_strength}. The classification accuracy decreases as the attack strength $\epsilon$ increases. As expected, when the perturbation is smaller, SODEF can provide good adversarial robustness.

\begin{figure}[!htb]
\centering
\includegraphics[width=0.55\textwidth]{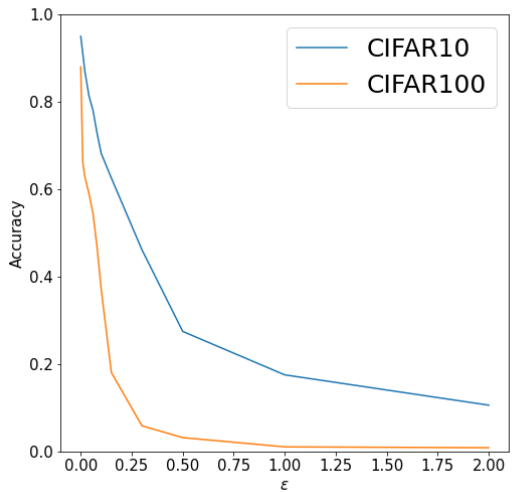}
\vspace{-0.2in}
\caption{Classification accuracy under FGSM attack with different parameters  $\epsilon$.}\label{fig:diff_strength}
\end{figure}

Since perturbation in the input $x$ is altered by the feature extractor $h_{\bphi}$, in another experiment, we also directly perturb the feature $\bz(0)$ instead of the input $x$. The results are shown in \cref{fig:eig_densec}, where we can observe how the accuracy changes with the attack strength $\epsilon$. Without any neural ODE layer, the classification accuracy decays dramatically  and becomes zero when $\epsilon$ is increased to $0.5$. However, while both ODE net and SODEF have demonstrate defense ability over the FGSM attack even when $\epsilon\ge 0.5$, SODEF performs much better than ODE, i.e., the classification accuracy at $\epsilon=0.5$ for SODEF is $59.9\%$ while ODE net achieves only $18.5\%$ accuracy.

\begin{figure}[!htb]
\centering
 \begin{subfigure}[t]{0.31\columnwidth}
        \centering
 \includegraphics[width=1\textwidth]{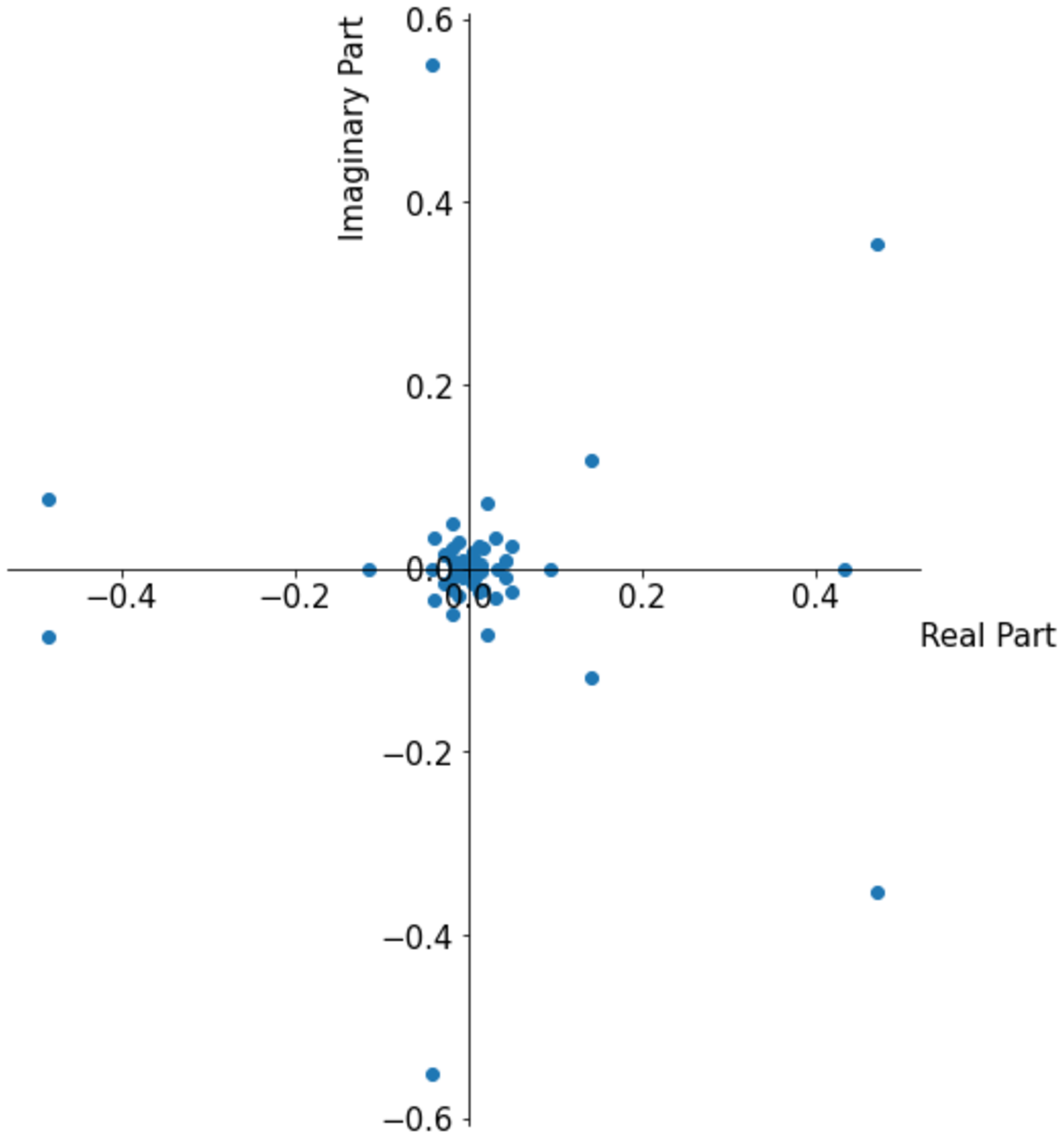}
        \caption{Eigenvalues of $\nabla f_{\btheta}(\bz(0))$ of ODE net.}
        \label{fig:eig_densea}
    \end{subfigure}
    ~
    \begin{subfigure}[t]{0.31\columnwidth}
        \centering
 \includegraphics[width=1\textwidth]{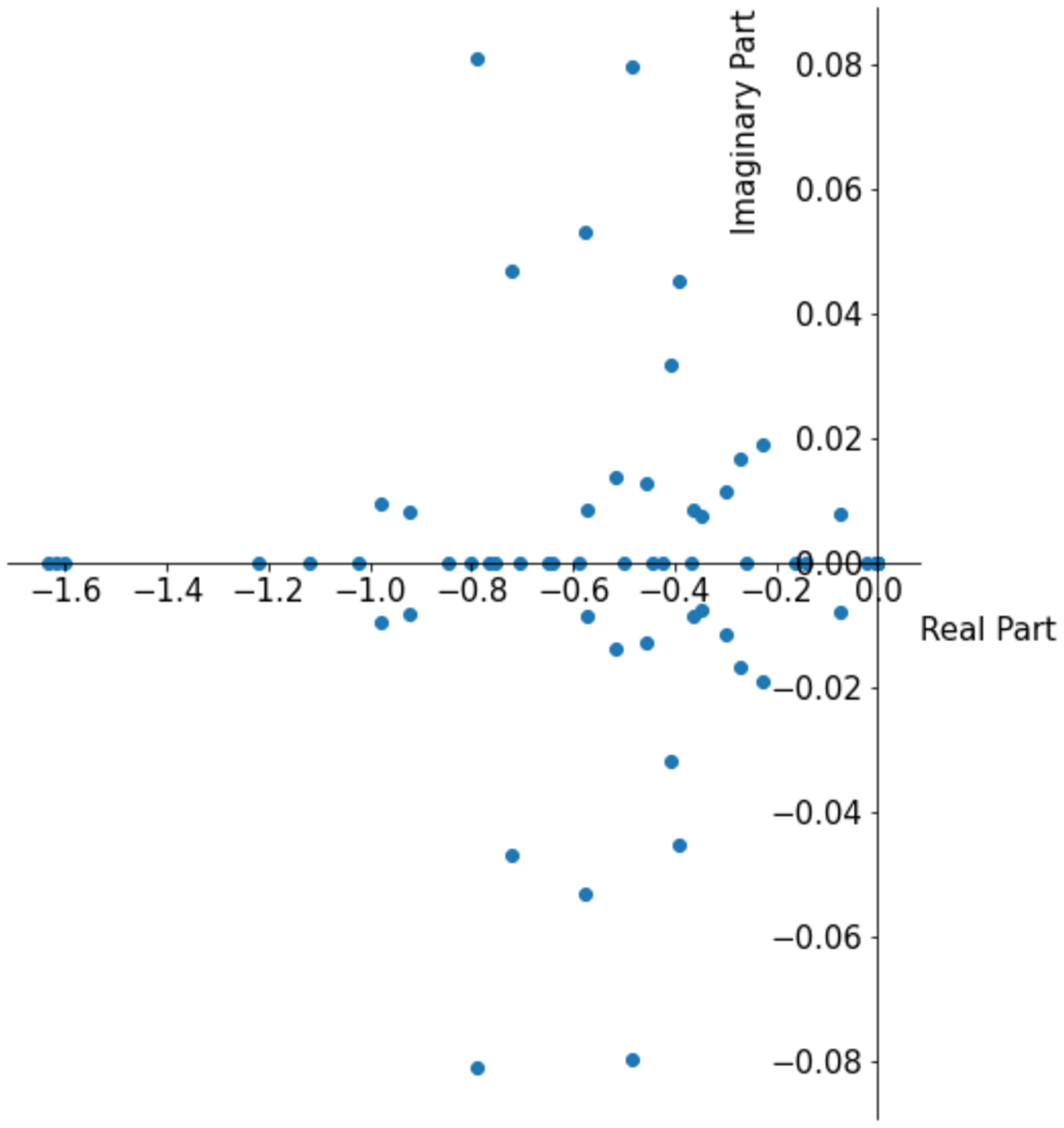}
        \caption{Eigenvalues of $\nabla f_{\btheta}(\bz(0))$ of SODEF.}
       \label{fig:eig_denseb}
    \end{subfigure}
    ~
        \begin{subfigure}[t]{0.31\columnwidth}
        \centering
     \includegraphics[width=1\textwidth]{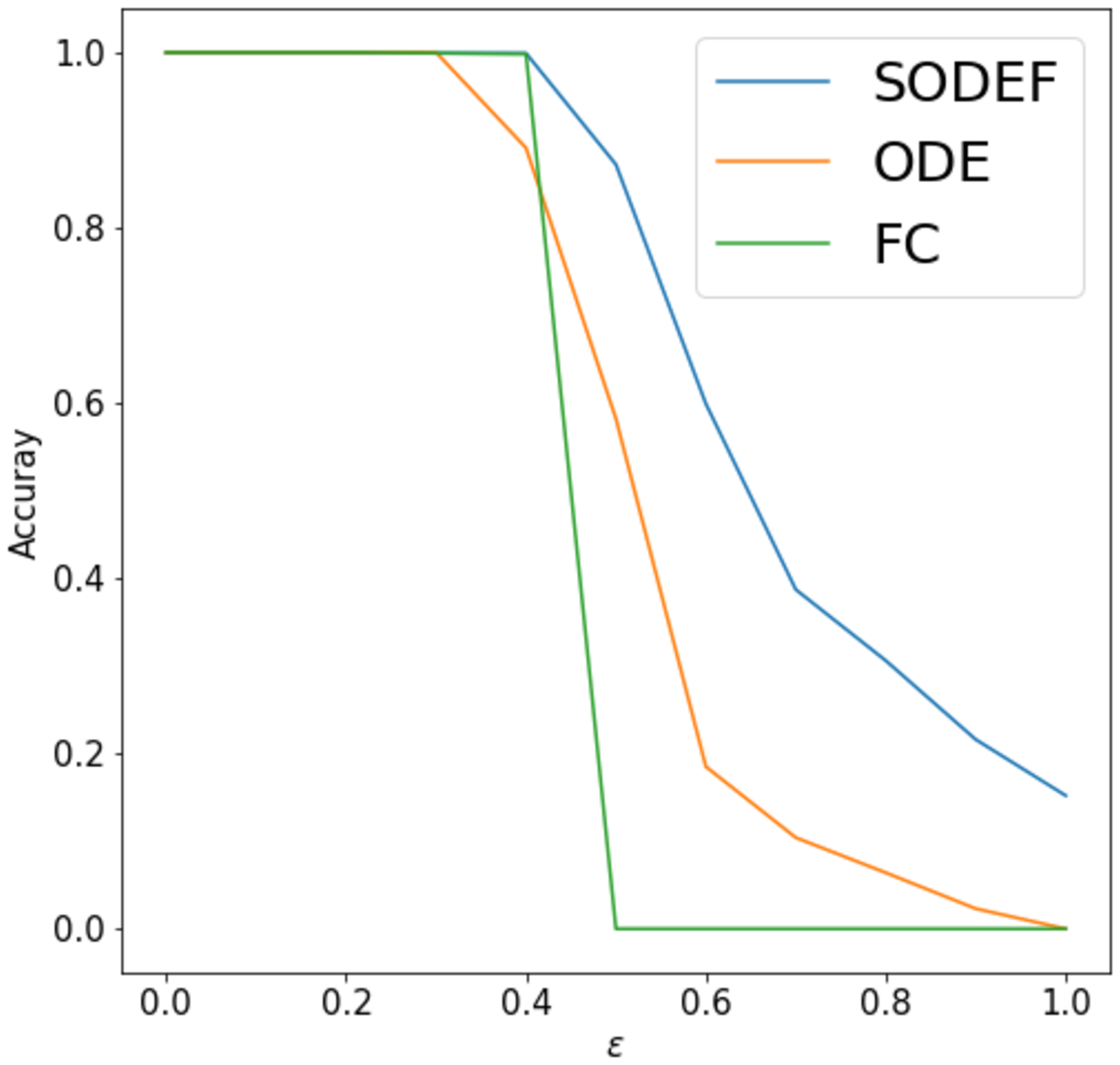}
        \caption{Classification accuracy under FGSM attack with different parameters $\epsilon$.}
       \label{fig:eig_densec}
    \end{subfigure}
  \caption{Eigenvalue visualization for the Jacobian matrix $\nabla f_{\btheta}(\bz(0))$ from ODE net and SODEF, and classification accuracy under FGSM attack with different parameters $\epsilon$.}
  \label{fig:eig_dense}
\end{figure}

\section{Eigenvalues Visualization}
To further show the effect of the SODEF regularizers in \cref{eq:obj_emp}, we compute the eigenvalues of $\nabla f_{\btheta}(\bz(0))$ from ODE net (no stability constraint is imposed) and SODEF, as shown in \cref{fig:eig_densea,fig:eig_denseb}. We observe that the eigenvalues of the Jacobian matrix from SODEF are located in the left plane, i.e., the eigenvalues all have negative real parts. As a comparison, a large number of ODE net's eigenvalues are located in the right plane, as shown in \cref{fig:eig_densea}. 

\section{Classification Accuracy on  CIFAR10  using AutoAttack}
In the paper, we study the influence of the SODEF integration time $T$ using CIFAR10. Here in this section, we further test the influence of the SODEF integration time $T$ using CIFAR10. We use the model  provided by  \cite{pytorchcifar10}, which has nearly $95\%$ clean accuracy on CIFAR10, as $h_{ \phi }$. We use  AutoAttack with  $\mathcal{L}_{2}$ norm ($\epsilon=0.5$). We still observe that for all the four individual attacks and the strongest ensemble AutoAttack, SODEF  has a good performance for large enough integration time $T$.
\begin{table}[!tbh]
\centering
\caption{Classification accuracy (\%) under AutoAttack  on adversarial CIFAR10 examples with  $\mathcal{L}_2$ norm, $\epsilon=0.5$ and different integration time $T$ for SODEF.} \label{tab:r3_1}
\begin{tabular}{ccccccccccc} 
\toprule
Attack / $T$       &1  &3  & 5   & 6 &7  &8  &9  & 10 \\
\midrule
%\midrule 
\multirow{1}{*}{Clean}  &  95.10 &  94.95      & 95.05             &  95.00    &   94.92  & 95.10  & 95.11 & 95.02  \\   
\multirow{1}{*}{APGD$_{\text{CE}}$}   &3.52 & 8.20      &8.59       &12.59      &73.05         &  92.41  &92.58   &\tb{92.67}  \\   
\multirow{1}{*}{APGD$^{\text{T}}_{\text{DLR}}$}  & 8.20& 8.94&9.49  & 11.58     &  71.06  &\tb{92.19}  &91.88   &91.80  \\      
\multirow{1}{*}{FAB$^{\text{T}}$}  & 26.95  & 93.93      &95.05   & 95.00      & 94.92   &95.10   &\tb{95.11} &95.02   \\    
\multirow{1}{*}{Squre}   &  73.95 & 76.95      &80.30             & 80.62      &  81.30  &80.86  &83.59 & \tb{85.55} \\   
\multirow{1}{*}{AutoAttack}   &  0.10 & 2.73      &3.34         & 4.52      & 24.22    &  80.16  &81.25  &\tb{82.81}   \\      
\bottomrule
\end{tabular}
\end{table}

\section{Classification Accuracy on CIFAR100 using FGSM and PGD}

We further test SODEF on the CIFAR100 dataset. \cref{tab:adv_cifar100_SODEF} shows that SODEF is consistenly more robust than all the competitors, especially when stronger attack is applied.
\begin{table}[!tbh]
\centering
\caption{Classification accuracy (\%) on adversarial CIFAR100 examples.} \label{tab:adv_cifar100_SODEF} 
\vspace{-6pt}
\small
\begin{tabular}{cccccccc} 
\toprule
Attack & Para.                          &no ode  &ODE  &  TisODE    & SODEF \\
\midrule
None  & -                               & 88.3      &88.1           &     88.14               &    88.0         \\ 
%\midrule       
\multirow{1}{*}{FGSM}  &  $\epsilon=0.1$ & 25.32      &29.50          &   31.90         &    {\bf37.67}      \\   
\multirow{1}{*}{PGD}  &  $\epsilon=0.1$ & 2.39      &3.36            &  6.82           &    {\bf 22.35}      \\   
%\midrule 
%\multirow{1}{*}{C\&W} &  $\kappa=1$    &  ?                  &        ?          &   {\bf ?}    \\ 
%\midrule    
%\multirow{1}{*}{BSA} & $\alpha=0.8$      & ?                  &     ?             & ?  \\ 
%\midrule
%\multirow{1}{*}{JSMA} & $\gamma=0.6$      & ?                &     ?             &?  \\ 
%\midrule
%\# params & -              &  818,334              &  401,168          & 490,209   \\
\bottomrule
\end{tabular}
\end{table}

We then do an experiment for the CIFAR100 dataset where we fix the entire feature extractor (EffcientNet, pretrained on imagenet and CIFAR100, used in our paper already) except for the last layer in the extractor. The results are shown as follows in \cref{tab:r2_1}. In the first line we use the pretrained EffcientNet with our SODEF and with fine-tuning. In the second line we use a fixed pretrained EffcientNet and with trainable SODEF. In the third line we just attack the pretrained EffcientNet (no ODE and no training since this opensource model has already been trained on CIFAR100 and get $88\%$ clean accuracy).

We observe the fine-tuned model performs the best among three methods and fixed (w/ SODEF) performs better than the vanilla model. One possible reason for fixed (w/ SODEF) being inferior to the fine-tuned model  is that the pretrained feature extractor does not have a well diversified $\bz(0)$.}

\begin{table}[!tbh]
\centering
\caption{Classification accuracy (\%) using SODEF under PGD and FGSM attack on adversarial CIFAR100 examples with parameters $\epsilon=0.1$.} 
\vspace{-6pt}\label{tab:r2_1}
\small
\begin{tabular}{cccccccc} 
\toprule
Model / Attack                      & clean     &FGSM &PGD  \\
\midrule
%\midrule       
\multirow{1}{*}{Fine-tuned (w/ SODEF)}  &88.0   &\tb{37.67} &\tb{22.35} \\
\multirow{1}{*}{Fixed (w/ SODEF)}     &87.1   &33.65  &17.32 \\ 
\multirow{1}{*}{Fixed (no ODE block)}  &88.0  &25.32 &2.39 \\

%\midrule 
%\multirow{1}{*}{C\&W} &  $\kappa=1$    &  ?                  &        ?          &   {\bf ?}    \\ 
%\midrule    
%\multirow{1}{*}{BSA} & $\alpha=0.8$      & ?                  &     ?             & ?  \\ 
%\midrule
%\multirow{1}{*}{JSMA} & $\gamma=0.6$      & ?                &     ?             &?  \\ 
%\midrule
%\# params & -              &  818,334              &  401,168          & 490,209   \\
\bottomrule
\end{tabular}
\end{table}

\section{Further Ablation Studies}
\label{sect:more_abla}

We performed more experiments on the CIFAR10 dataset to better understand the proposed SODEF. We let the feature extractor $h_\phi$ be  ResNet18 (which ends up with a feature vector $\bz(0)\in\Real^{10}$) and let $f_\theta$ be a linear function such that $f_\theta(\bz(t))=-\bC\bC\T\bz(t)$ where $\theta=\bC\in\Real^{10\times 10}$. Since $-\bC\bC\T$ is a negative definite matrix for full rank $\bC$, the ODE block is guaranteed to be asymptotically stable according to \cref{thm:stab_lin}. Note that this setup is a special case of our proposed SODEF architecture, which allows $f_\theta$ to be non-linear. The final FC is fixed to be $\bV$ given in \cref{cor:reg_simplex}. For comparison, we construct three other neutral network architectures as described in \cref{tab:ablation}.
\begin{table}[!tbh]
\centering
\caption{Architectures used in our ablation study.} \label{tab:ablation} 
\vspace{-6pt}
\small
\begin{tabular}{cccc} 
\toprule
Arch. & $h_\phi$ & $f_\theta$ & FC \\
\midrule
SODEF & ResNet18 & $-\bC\bC\T\bz(t)$ & $\bV$        \\ 
%\midrule       
Net1 & ResNet18 & $-\bC\bC\T\bz(t)$ & random orth. matrix        \\ 
%\midrule    
Net2 & ResNet18 & $-\bC\bC\T\bz(t)$ & linear (trainable)        \\ 
%\midrule    
Net3 & ResNet18 & $\bC\T\bz(t)+{\bf b}$ & $\bV$        \\ 
\bottomrule
\end{tabular}
\end{table}

\cref{tab:ablation_cifar10} provides the following insights: 1) Comparing SODEF with Net3 highlights the importance of stability of the ODE block; 2) Comparing SODEF with Net 1 and Net2 shows the superiority of our designed FC layer; 3) By comparing Net2 with Net1 and SODEF, we see that enforcing $\bz(T)$ to align with a prescribed feature vector, e.g., a row in $\bV$, improves model robustness.

\begin{table}[!tbh]
\centering
\caption{Classification accuracy (\%) on adversarial CIFAR10 examples. We apply PGD for 20 iterations to generate adversarial examples.} \label{tab:ablation_cifar10} 
\vspace{-6pt}
\small
\begin{tabular}{cccccc} 
\toprule
Attack & Para.             & SODEF     & Net1  &  Net2   & Net3 \\
\midrule
None  & -                  & 91.5      & 91.1  &  91.3   & {\bf 91.9}         \\ 
PGD  &  $\epsilon=0.1$     & {\bf 29.2}      & 24.7  &  5.6    & 18.8      \\   
\bottomrule
\end{tabular}
\end{table}

\section{Further Transferability Studies}
We use the same SODEF setup as in \cref{sect:more_abla}. In order to determine if SODEF's robustness is a result of gradient masking, we generate adversarial examples from two substitute networks whose architectures are given in \cref{tab:transfer} and feed the generated examples into SODEF.

\begin{table}[!tbh]
\centering
\caption{Two substitute networks used to generate adversarial examples by running PGD attack with $\epsilon=0.1$ for 20 iterations. Both networks have no ODE block, i.e., $f_\theta=0$} \label{tab:transfer} 
\vspace{-6pt}
\small
\begin{tabular}{cccc} 
\toprule
Arch. & $h_\phi$ & $f_\theta$ & FC \\
\midrule
NetA & ResNet18 & - & linear (trainable)        \\ 
%\midrule    
NetB & ResNet18 & - & $\bV$       \\ 
\bottomrule
\end{tabular}
\end{table}

\cref{tab:transfer_cifar10} provides the following insights: 1) The SODEF column shows that performing the white-box attack directly on SODEF is more effective than performing transferred attacks using substitute models such as NetA and NetB and thereby implying that the ODE block in SODEF does not mask the gradient when performing PGD attack. 2) The NetA and NetB columns show that even when there is no ODE block, NetB using $\bV$ as the last FC layer is much more robust than NetA whose FC layer is a trainable dense layer.

\begin{table}[!tbh]
\centering
\caption{Classification accuracy (\%) on adversarial CIFAR10 examples. We generate adversarial examples by attacking NetA and NetB using PGD method (run for 20 iterations with $\epsilon=0.1$) and then feed them to SODEF.} \label{tab:transfer_cifar10} 
\vspace{-6pt}
\small
\begin{tabular}{c|ccc} 
\toprule
Adv. examples gen. by:  & SODEF     & NetA  &  NetB   \\
\hline
 No Attack                        & 91.5     & 93.6        & 91.5        \\ 
 SODEF  & 29.2     & -         & -         \\   
 NetA  & 43.9     & 2.6         & -         \\   
 NetB  & 51.3     & -           & 20.5     \\   
\bottomrule
\end{tabular}
\end{table}
{
\section{Different ODE Solvers}
 We test different ODE solvers including Runge-Kutta of order 5, Runge-Kutta of order 2, Euler method, Midpoint method, and Fourth-order Runge-Kutta with 3/8 rule. All the ODE solvers tested in \cref{tab:r2_2} show similar performance. For Euler method, Midpoint method, and Fourth-order Runge-Kutta with 3/8 rule, we set step size to $0.05$. See \url{https://github.com/rtqichen/torchdiffeq}.

\begin{table}[!tbh]
\centering
\caption{Classification accuracy (\%) using SODEF under FGSM attack on  adversarial CIFAR10 examples with parameters $\epsilon=0.1$ and different ODE solvers.} \label{tab:r2_2}
\vspace{-6pt}
\small
\begin{tabular}{cccccccc} 
\toprule
Model / Attack                      & clean     &FGSM  \\
\midrule
%\midrule       
\multirow{1}{*}{Runge-Kutta of order 5} &95.00  &68.05 \\
\multirow{1}{*}{ Runge-Kutta of order 2} &95.12  &69.95 \\ 
\multirow{1}{*}{ Euler method}         &95.20  &70.22\\ 
\multirow{1}{*}{Midpoint method}        &95.10  &70.86\\ 
\multirow{1}{*}{Fourth-order Runge-Kutta with 3/8 rule} &95.07  &69.50 \\ 

%\midrule 
%\multirow{1}{*}{C\&W} &  $\kappa=1$    &  ?                  &        ?          &   {\bf ?}    \\ 
%\midrule    
%\multirow{1}{*}{BSA} & $\alpha=0.8$      & ?                  &     ?             & ?  \\ 
%\midrule
%\multirow{1}{*}{JSMA} & $\gamma=0.6$      & ?                &     ?             &?  \\ 
%\midrule
%\# params & -              &  818,334              &  401,168          & 490,209   \\
\bottomrule
\end{tabular}
\end{table}
}

\section{Proofs of Results in Paper}

In this section, we provide detailed proofs of the results stated in the main text.

\subsection{Proof of \cref{thm:reg_simplex}}
\label{sect:proof_thm:reg_simplex}
For simplicity, let $a=a(\bv_1,\ldots,\bv_k)$. Since $0\leq \|\sum_{i=1}^k \bv_i\|^2=\sum_i \|\bv_i\|^2 + 2\sum_{i\leq j}\bv_i\T\bv_j\leq k + 2\sum_{i\leq j}a=k+k(k-1)a$, we have $a\geq 1/(1-k)$. We next prove $a=1/(1-k)$ is attainable. We can construct a  $k\times k$ matrix $\bB$ with $\bB(i,i)=1,\forall i$ and $\bB(i,j)=1/(1-k),\ \forall i\neq j$. It can be verified that $\bB$ is a diagonally dominant matrix. Due to the fact that a symmetric diagonally dominant real matrix with nonnegative diagonal entries is positive semi-definite and hence a Gram matrix \gls{wrt} the Euclidean inner product \cite{horn2012matrix}, there exists a set of vectors $\bv_i$ such that $\bv_i\T\bv_j=\bB(i,j)=1/(1-k)$ for $i\ne j$. In this case, $a(\bv_1,\ldots,\bv_k)=1/(1-k)$ and the proof is complete.

\subsection{Proof of \cref{cor:reg_simplex}}
\label{sect:proof_cor:reg_simplex}
We may decompose $\bB$ as 
\begin{align*}
        \bB&=(\bSigma^{1/2}\bU\T)\T(\bSigma^{1/2}\bU\T) \\
    &=\left(\bQ\bSigma^{1/2}\bU\T\right)\T \left(\bQ\bSigma^{1/2}\bU\T\right),
\end{align*}
and hence $\bB$ is the Gram matrix of $\set{\bv_1,\ldots,\bv_k}$. The result follows from the construction of $\bB$.

\subsection{Proof of \cref{lem:exist_f}}
\label{sect:proof_lem:exist_f}
% \red{[Does your $\bA_i$ have to be non-singular?][Suggest to skip whole proof and reference the Whitney extension theorem.]}{[no need to be non-singular, okay, let's skip the whole proof, I first tried to prove this by interpolation and then realize that there exists this Whitney extension theorem.]}

% For the $n=1$ case, let polynomial $f(z) = \prod_{i=1}^{k} (z-z_i)p(z)$, where $p(z)=\sum_{i=0}^{k-1}b_iz^i$ is a $k-1$ order polynomial with $b_i\in \Real$ being the $k$ unknowns. Obviously $f(z)$ meets the requirements that $z_i, i= 1,...,k$ are its zeros. We next determine the $k$ unknowns using the $k$ derivatives.
% \begin{align}
%     f'(z) = \sum_{i=1}^{k} \prod_{j\ne i}(z-z_j)p(z) + p'(z)\prod_{i=1}^{k} (z-z_i)\label{eq:n1case}
% \end{align}
% Substituting $z_i$ to \cref{eq:n1case}, we get $f'(z_i) = p(z_i)\prod_{j\ne i}(z_i-z_j) = A_i$ (note here $A_i$ is a scalar). Since $z_i\ne z_j$,  the $k$ unknowns with $k$ equations can be resolved using Vandermonde matrices or Lagrange interpolation \cite{horn2012matrix}. 

% For the general $n>1$ case, %it is not obvious to use a polynomial interpolation to achieve the conclusion as above $n=1$ case.
% we instead prove it abstractly. 

The set of finite points $\{\bz_1,...,\bz_k\}$ is closed and this lemma is an immediate consequence of the Whitney extension theorem \cite{mcshane1934extension}.

\subsection{Proof of \cref{lem:noexist_f}}\label{sect:proof_lem:noexist_f}  

If a continuous function $f$ is such that $f(\bz) =0$ $\nu_{\bphi}$-almost surely, then $f(\bz)=0$ for all $\bz\in E'$. Suppose otherwise, then there exists $\bz_0$ such that $\norm{f(\bz_0)}_2>0$. From continuity, there exists an open ball $B\in E'$ containing $\bz_0$ such that $\norm{f(\bz)}_2>0$ for all $\bz\in B$. Then since $B$ has non-zero Lebesgue measure, $\nu_{\bphi}(B)>0$, a contradiction. Therefore, every $\bz\in E'$ has $\nabla f(\bz)=0$ and cannot be a Lyapunov-stable equilibrium point.

\subsection{Proof of \cref{thm:exist_f}}\label{sect:proof_thm:exist_f} 

%\begin{Theorem_4}
%Suppose \cref{assumpt:input,assumpt:disjoint}. The function space satisfying the constraints in %\cref{eq:re1,eq:re2,eq:re3} is non-empty for all $\epsilon>0$ and there exist functions in this space such %that each support $E_l$ contains at least one Lyapunov-stable equilibrium point.
%\end{Theorem_4}

Consider $f(\bz) = [f^{(1)}(\bz^{(1)}),...,f^{(n)}(\bz^{(n)})]$ with each $f^{(i)}(\bz^{(i)})\in C^1(\Real,\Real)$. Since  $f^{(i)}(\bz^{(i)})$ only depends on $\bz^{(i)}$, $\nabla f_{\btheta}(\bz)$ is a diagonal matrix with all off-diagonal elements being $0$. The constraint \cref{eq:re3} is thus satisfied immediately and it suffices to show that there exists such a $f$ satisfying the constraints \cref{eq:re1,eq:re2}.  

Select a $\bz_l=(\bz\tc{1}_l,\ldots,\bz\tc{n}_l)$ from the interior of each $E_l$, $l=1,\dots,L$. Let  $f^{(i)}(\bz^{(i)})=-\beta(\bz^{(i)} - \bz_l^{(i)})$ on each $E_l$, where $\beta>0$. Then $f(\bz)$ satisfies \cref{eq:re2} for all $\beta>0$ and $\bz_l$ is a  Lyapunov-stable equilibrium point for each $l$ since $\nabla f_{\btheta}(\bz_l)$ is a diagonal matrix with negative diagonal values. Since each $E_l \subset\Real^n$ is compact, we have that $\forall \epsilon>0$, $\exists \beta>0$ sufficiently small such that $ |f^{(i)}(\bz^{(i)})|<\epsilon$ for all $\bz \in \bigcup_l E_l $. The constraint \cref{eq:re1} is therefore satisfied for $f(\bz)$ with a sufficiently small $\beta$. Since $\bigcup_l E_l $ is closed, the Whitney extension theorem \cite{mcshane1934extension} can be applied to extend $f(\bz)$ to a function in $C^1(\Real^n,\Real^n)$.

\bibliographystyle{IEEEtran}
\bibliography{IEEEabrv,StringDefinitions,adv_dnn}

\end{document}